\pdfoutput=1
\documentclass[11pt]{article}

\newcommand{\cmark}{\textcolor{green}{\checkmark}}
\newcommand{\xmark}{\textcolor{red}{\ding{55}}}

\usepackage{booktabs}
\usepackage{multirow}
\usepackage{graphicx}
\usepackage{tabularx}
\usepackage{amssymb}
\usepackage{pifont}
\usepackage{float}
\usepackage{enumitem}
\usepackage{amsmath}
\setlist[itemize]{leftmargin=*}

\usepackage[preprint]{acl}

\usepackage{times}
\usepackage{latexsym}

\usepackage[T1]{fontenc}
\usepackage[utf8]{inputenc}

\usepackage{microtype}

\usepackage{inconsolata}

\usepackage{graphicx}

\title{Measuring What Makes You Unique: Difference-Aware User Modeling for Enhancing LLM Personalization}

\newcommand{\meanstdllm}[2]{$\scalebox{0.82}{#1}_{\pm \scalebox{0.72}{#2}}$}

\author{
 \textbf{Yilun Qiu\textsuperscript{1\thanks{Equal contribution}}},
 \textbf{Xiaoyan Zhao\textsuperscript{2\footnotemark[1]}},
 \textbf{Yang Zhang\textsuperscript{1\thanks{Corresponding Author}}},
 \textbf{Yimeng Bai\textsuperscript{3}},
\\
 \textbf{Wenjie Wang\textsuperscript{3}},
 \textbf{Hong Cheng\textsuperscript{2}},
 \textbf{Fuli Feng\textsuperscript{3}},
 \textbf{Tat-Seng Chua\textsuperscript{1}}
\\
 \textsuperscript{1}National University of Singapore \\
 \textsuperscript{2}The Chinese University of Hong Kong \\
 \textsuperscript{3}University of Science and Technology of China \\
 \small
 qiuyilun@u.nus.edu, xzhao@se.cuhk.edu.hk, 
 zyang1580@gmail.com,
 baiyimeng@mail.ustc.edu.cn, 
 \\
 \small
 wenjiewang96@gmail.com, hcheng@se.cuhk.edu.hk, fulifeng93@gmail.com, dcscts@nus.edu.sg
}

\begin{document}
\maketitle
\begin{abstract}

Personalizing Large Language Models (LLMs) has become a critical step in facilitating their widespread application to enhance individual life experiences. In pursuit of personalization, distilling key preference information from an individual's historical data as instructional preference context to customize LLM generation has emerged as a promising direction. However, these methods face a fundamental limitation by overlooking the inter-user comparative analysis, which is essential for identifying the inter-user differences that truly shape preferences. To address this limitation, we propose Difference-aware Personalization Learning (DPL), a novel approach that emphasizes extracting inter-user differences to enhance LLM personalization. DPL strategically selects representative users for comparison and establishes a structured standard to extract meaningful, task-relevant differences for customizing LLM generation. Extensive experiments on real-world datasets demonstrate that DPL significantly enhances LLM personalization. 
We release our code at \url{https://github.com/SnowCharmQ/DPL}.
% We release our code on Github\footnote{\href{https://github.com/SnowCharmQ/DPL}{https://github.com/SnowCharmQ/DPL}}.

\end{abstract}

\section{Introduction}
With continuous efforts, Large Language Models (LLMs)~\cite{gpt4,zhao2024pacar,llama3,deepseek} have demonstrated increasingly higher levels of intelligence, sparking an unprecedented enthusiasm for applying them to individuals' daily lives to enhance personal thinking, planning, and overall life experiences~\cite{tallrec,personalAgent,zhao2025exploring,collm}. However, the ``one-size-fits-all'' paradigm of general LLMs becomes ineffective when serving individual users, as users have distinct preferences and 
% may expect different outputs even for the same request. 
would expect personalized generation from LLMs~\cite{userinterestjourney}.
This motivates an exciting new direction: \textit{LLM personalization}~\cite{persoSurvey1,persoSurvey2,persoSurvey3,persoSurvey4} -- adapting LLMs to generate text responses that align with each individual's unique preferences.
% aligning individual user preferences with the generation capabilities of LLMs to deliver more user-centric content, thus deeply and meaningfully serving each individual.

% Towards LLM personalization, the core lies in injecting user-personalized information into LLMs and leveraging them for customizing LLMs' outputs. 
% The lightweight "memorize-then-inject" framework offers a promising solution by memorizing users’ historical data and then extracting key information as the prompting contexts for customizing LLM output. According to the extraction method utilized, existing methods can be generally categorized into: 1) retrieve-based method, which just retrieves the parts relevant to the current request from the user's history for LLM use; 2) summarization-based method, which further summarizes user history to distill preference for LLM use. 

% Towards LLM personalization, 
LLM personalization fundamentally relies on incorporating user-specific information into LLMs to tailor their generation~\cite{zhao2025nextquill,pad,rlpa}.
The ``memorize-then-inject" framework offers a promising and economical solution -- storing users' historical data and then extracting key information to serve as instructional contexts for customizing LLM responses~\cite{lamp,automr,longmemeval}. Depending on the extraction method employed, existing approaches can generally be categorized into: 1) Retrieval-based methods~\cite{teachllm,pearl}, which just retrieve the parts of the user's history relevant to the current request for LLM use, and 2) Summarization-based methods~\cite{summ, recsumm}, which further condense the user history to distill preferences for LLM use.

While these methods~\cite{pearl,recsumm} have made notable progress, we contend that they share a common limitation: 
they overlook the inter-user comparative analysis, which is essential for identifying the differences that truly shape preferences.
% they overlook the personalized information extracted from the inter-user comparative analysis, thus limiting personalization abilities. 
As acknowledged in psychology and behavioral science~\cite{snyder1977abnormality,snyder2012uniqueness,irmak2010you},
it is the differences among individuals that make them unique, while their uniqueness shapes their preferences. This underscores that effective personalization hinges on identifying and understanding these differences. 
As a result, without considering inter-user comparisons to extract differences, current methods would be suboptimal for personalization.
In this work, we propose difference-aware user modeling -- extracting the differences among users as key preference information -- to enhance LLM personalization. 
% To extract the differences, 
To achieve this, distinguishing each user from others and then summarizing the differences (using an LLM) to serve as the instructional contexts for personalization provides an intuitive solution. However, it faces two challenges: 1) comparing to all users may be impractical due to the high cost and complexity of extracting differences across all others; and 2) without proper guidance, the differences extracted by the LLM may be irrelevant or meaningless for the task. Therefore, we must address: 1) how to select appropriate users for comparison, and 2) how to extract task-relevant differences during comparison effectively.

To address these challenges, we introduce a novel \textit{Difference-aware Personalization Learning} (DPL) method. DPL selectively chooses representative users for comparison and establishes an extraction standard to effectively capture task-relevant differences, using the extracted differences as the LLM’s instructional contexts for personalization. 
% Specifically, we cluster users with similar data and compare only with group center users, facilitating comparison while maintaining a global perspective. 
Specifically, we cluster users based on data similarity and select only cluster-center users for comparison, ensuring efficiency while maintaining a global perspective.
% To extract task-relevant differences, we identify key preference dimensions -- writing style, emotional tone, and semantic style -- based on existing literature. We then guide the difference extraction model (an LLM) to focus on these dimensions to extract the relevant differences.
% \textcolor{blue}{To extract task-relevant differences, we distill some key preference dimensions (\textit{e.g.}, writing style, emotional style, and semantic style) from existing literature.}
% We then guide the difference extraction model (an LLM) to focus on the dimensions to extract the relevant differences.
To extract task-relevant differences, we structure the difference extraction model (an LLM) to focus on key predefined dimensions for capturing differences, rather than operating freely. Three key preference dimensions are distilled from existing literature as examples: writing style, emotional tone, and semantic content.
% Three key preference dimensions are distilled from existing literature as examples, including writing style, emotional tone, and semantic style.

% To extract task-relevant differences, we first identify key preference aspects for personalized text generation—writing style, emotional tone, and semantic style—based on a review of existing literature. We then guide the model to focus on these aspects to ensure the extraction of task-relevant differences.
% We validate our method on the personalized review generation task, a representative personalized text generation benchmark, demonstrating that DPL outperforms state-of-the-art methods.
We validate our DPL method on a representative personalized text generation task, namely, review generation~\cite{amazon20181, reviewllm,pgraphrag}. Extensive results demonstrate that DPL effectively enhances LLM personalization.
% that DPL outperforms state-of-the-art approaches.

% Extensive experiments demonstrate that DPL outperforms the state-of-art methods across multiple evaluation metrics. We release our benchmark\footnote{Huggingface} and codes\footnote{GitHub}.

% \xyz{

% The contributions of our paper include:
The main contributions of this work are summarized as follows:
% \vspace{-0.4em}
\begin{itemize}
    \setlength{\itemsep}{0pt}
    % \item For the first time, we propose Personalized Difference Learning for distinguishing the personalized information in personalized text generation.

    \item We emphasize the philosophy that ``differences make us unique'' as the key to personalization and propose using difference-aware user modeling to enhance LLM personalization.
    
    \item 
    % We propose a novel Difference-aware Personalization Learning method for LLM, incorporating a selective comparison mechanism and a difference extraction standard to achieve effective difference extraction.
    We propose a novel Difference-aware Personalization Learning method for LLM, incorporating a selective comparison mechanism and structured difference extraction mechanism to achieve effective difference extraction.

    % \item We establish a personalized text difference extraction standard for guiding DPL in identifying and preserving task-relevance differences.
    \item Extensive experiments on real-world datasets demonstrate that DPL achieves state-of-the-art performance, highlighting its effectiveness in personalized text generation.
\end{itemize}
\begin{figure*}[t]
    % \vspace{-0.5em}
    \centering
    \includegraphics[width=0.85\linewidth]{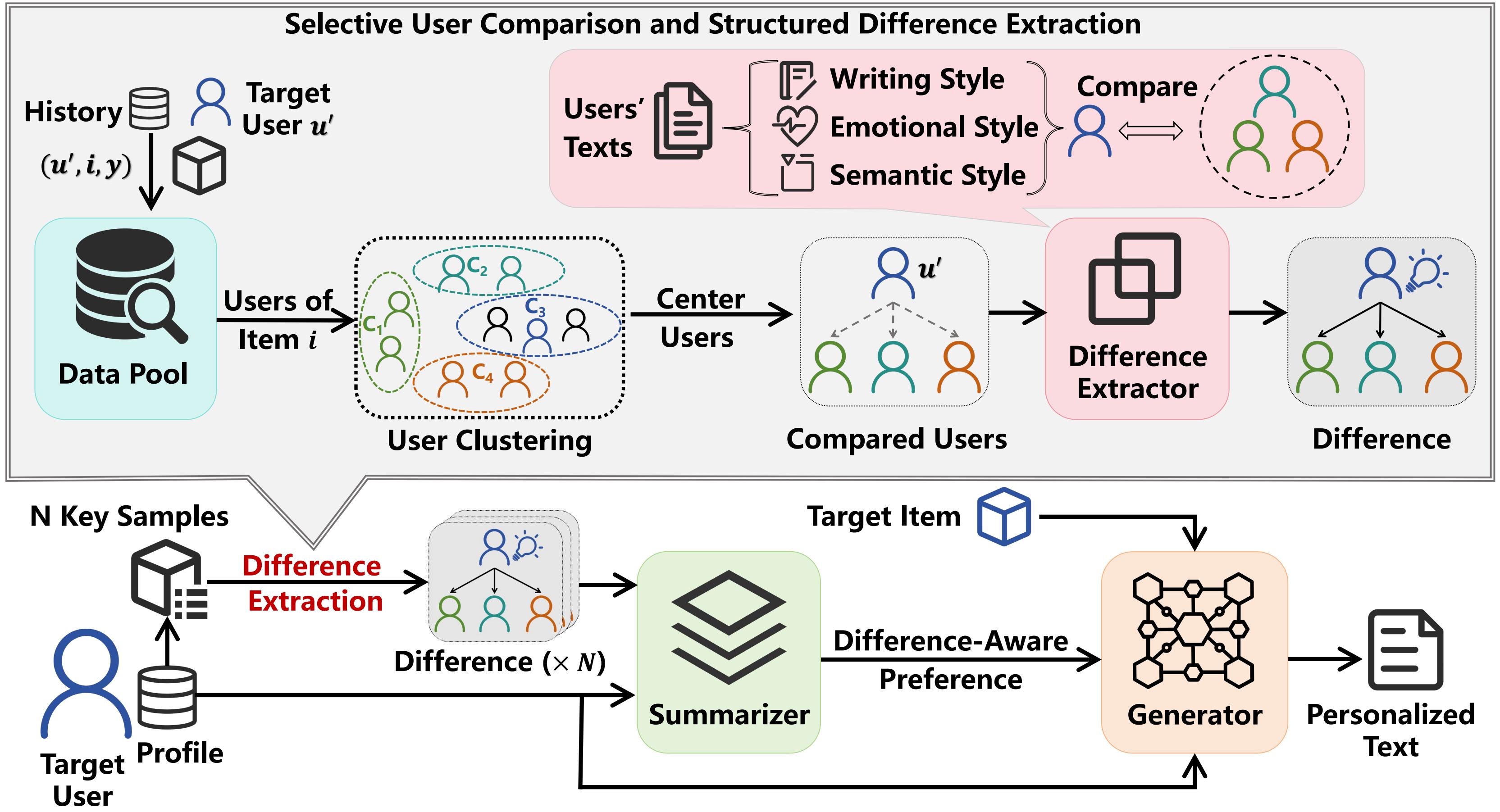}
    \caption{Overview of the proposed DPL method, which enables difference-aware preference extraction for LLM personalization. It extracts differences through selective user comparison via clustering and structured difference extraction along fixed dimensions (writing style, emotional style, and semantic style).}
    \label{main_method}
\end{figure*}

\section{Preliminary}

\noindent\textbf{Problem Formulation.}
This work focuses on personalizing LLMs to generate text outputs that align with user preferences, \textit{i.e.,} achieving personalized text generation. We assume that users have a set of historical texts, either written or preferred by them, reflecting the users' preferences.
We could leverage this data to customize LLM's generation to meet user needs. 
Formally, let $\mathcal{D}$ denote the historical data for all users. Each sample in $\mathcal{D}$ is represented as $(u, i, y) \in \mathcal{D}$, where $u$ represents the user, $i$ denotes the item (or object) the user is focused on, and $y$ indicates the text written or preferred by $u$ for item $i$. 
% For a user $u$, all corresponding historical data is denoted as $\mathcal{D}_{u}$, and similarly, all historical data related to an item $i$ is denoted as $\mathcal{D}_{i}$.
When a target user $u^{\prime}$ submits a new request for text generation on a target item $i^{\prime}$, the LLM needs to generate text that satisfies the user’s preferences based on $\mathcal{D}$. 

% A typical solution is that, for each user $u$, extracting key information from the user's own historical data $\mathcal{D}_{u}$ via retrieval or summarization, and then treating the results as the preference context in instructions to guide LLM's generation. However, we believe that solely focusing $\mathcal{D}_u$ fails to extract the difference among users 

% Toward LLM personalization, for each user $u$, existing works usually focusing on $\mathcal{D}_{u}$ . However, 

Notably, without loss of generality, this paper primarily focuses on review generation, a representative personalized text generation task. In this context, the goal is to enable LLMs to generate personalized reviews for items such as movies or products, ensuring alignment with true user reviews.

\vspace{+5pt}
\noindent\textbf{Memorize-then-Inject Solution}. A typical approach in existing works for customizing LLM generation is to store the user's historical data and then extract key information from the data to distill preference information through retrieval or summarization when needed. This extracted information serves as a preference context in the instruction to guide the model’s generation. Formally, to generate personalized text for the focused user $u^{\prime}$ on the target item $i^{\prime}$, the method can be represented as follows:
\begin{equation}
    \hat{y} = \text{LLM}(u^{\prime},i^{\prime},\phi_{key}(\mathcal{D}_{u^\prime})),
\end{equation}
where $\hat{y}$ denotes the generated result, $\mathcal{D}_{u^\prime}$ denotes the historical data for $u^{\prime}$, and $\phi_{key}(\mathcal{D}_{u^\prime})$ represents the key information extraction process from $\mathcal{D}_{u^\prime}$. 
Notably, the extraction process usually also involves the utilization of an LLM.
This method does not require LLM retraining and can potentially manage lifelong historical data in a memory manner~\cite{LLMmemory}, making it a promising and cost-effective solution.

\section{Methodology}

In this section, we introduce the proposed \textit{Difference-Aware Personalization Learning} (DPL) method, beginning with its motivation and the overall framework, followed by a detailed explanation of each component.

\subsection{Overview}
Given that the differences define individual uniqueness, we argue that inter-user differences play a crucial role in preference learning for LLM personalization. 
% However, existing approaches rely solely on individual data for preference modeling, 
However, existing approaches overlook the inter-user comparisons for preference modeling, limiting the ability to extract meaningful differences for personalization. To address this limitation, we propose the DPL method. 
As shown in Figure~\ref{main_method}, DPL introduces an inter-user comparison mechanism to identify differences between users, which are then combined with the user's own key history to form the preference context in instructions, guiding LLM generation. This process can be formulated as:  
\begin{equation}\label{eq:generate}
    \hat{y} = \text{LLM}(u^\prime, i^\prime, \phi(\mathcal{D}_{u^\prime}; \mathcal{D})), 
\end{equation}  
where $\phi(\mathcal{D}_{u^\prime}; \mathcal{D})$ represents the difference-aware preference extracted by comparing a user's historical data $\mathcal{D}_{u^\prime}$ with other users' data in $\mathcal{D}$.

\vspace{+3pt}
\textbf{Difference-Aware Preference Extraction Process.} As illustrated in the bottom part of Figure~\ref{main_method}, for user \( u^{\prime} \), 
we assume that $\mathcal{D}_{u^\prime}$ contains $N$ key elements, denoted as $\mathcal{D}_{u^\prime}^{\star}$,
which can be obtained via retrieval~\cite{bm25}.
Then we would compare each element with those of some other users to extract differences and then summarize these differences across all samples and the key historical data to obtain the final preference. 
Specifically, for each element $(u^\prime, i, y) \in \mathcal{D}_{u^\prime}^{\star}$, we compare it with the texts of other users on the same item $i$, represented by $\mathcal{D}_{i}$, using a \textbf{selective user comparison and structured difference extraction} process. This process consists of two key steps, as shown in the upper part of the figure:
\vspace{-0.3em}
\begin{itemize}  
    \setlength{\itemsep}{0pt}
    \item \textit{\textbf{Selective user comparison}}: Instead of comparing with all users in $\mathcal{D}_{i}$, we selectively compare with a subset of representative users to reduce complexity and improve efficiency.  

    \item \textit{\textbf{Structured difference extraction}}: 
    % \qyl{We extract differences based on a predefined standard to ensure task relevance and meaningful personalization.}
    Differences are extracted according to predefined standards to ensure task relevance and meaningfulness.
    % Differences are extracted following a predefined standard to ensure 
    % relevance to the task and meaningful personalization.  
\end{itemize}  
\vspace{-0.3em}
The details of these steps will be elaborated later. After extracting differences for all $N$ elements, we summarize them as well as the key history $\mathcal{D}_{u^\prime}^{\star}$ with an LLM to obtain the final difference-aware representation for personalization. Formally, 
\begin{equation}\label{eq:sum}
\small
    \phi(\mathcal{D}_{u^\prime}; \mathcal{D}) = \text{LLM}_{\text{sum}}(\mathcal{D}_{u^\prime}^{\star};[d_{1},\dots,d_{n},\dots,d_{N}]),  
\end{equation}  
where $\text{LLM}_\text{sum}$ denotes the summarizer, and $d_{n}$ represents the extracted difference for the $n$-th element in $\mathcal{D}_{u\prime}^{*}$. 
% Notably, we also consider the user's own data to keep the preference information that can not described by the differences.
% Besides, the key history $\mathcal{D}_{u^\prime}^\star$ can be obtained via retrieval.
Notably, we also take into account the user's own data to retain preference information that cannot be captured by the differences, \textit{i.e.}, using the key history $\mathcal{D}_{u^\prime}^\star$ as additional inputs.
Next, we elaborate on the selective user comparison and structured difference extraction methods to explain how each $d_{n}$ is obtained.

\subsection{Selective User Comparison and Structured Difference Extraction}
When dealing with each important historical data sample \((u^\prime, i, y) \in \mathcal{D}_{u^\prime}^{\star}\), a straightforward approach for difference extraction is to have the LLM compare all users and summarize the differences. However, comparing all users is complex and inefficient. To address this, we selectively compare only representative users. Additionally, since differences can span multiple dimensions, allowing LLMs to extract them freely may introduce task-irrelevant information. To mitigate this, we introduce a structured difference extraction approach with a predefined standard to guide the extraction of meaningful and task-relevant differences.

% we 
% perform a selective user comparison to reduce the comparison difficulties and enhance efficiency, and a structured difference extraction to ensure extracting task-relevant difference. 

\subsubsection{Selective User Comparison}\label{sec:suc}
To select representative users for comparison, the core consideration is that they could represent the total population, making the comparison still provide a global view. To achieve this, we consider clustering the users based on their text and then selecting the user in the cluster center for comparison. Specifically, when dealing with each key element in history $(u^\prime, i, y)\in\mathcal{D}_{u^\prime}^\star$ for the target $u^{\prime}$, we just consider the users having texts on $i$, and then cluster these users into $K+1$ groups using the K-means algorithm~\cite{kmeans} according to the texts corresponding to them, obtaining the most central users for each group. We then select the $K$ central users that do not belong to groups of the target user as the users for comparison.
Formally,

\vspace{-4pt}
\begin{equation}\label{eq:cluster}
    u_{1}, \dots, u_{K} = \text{cluster}\_\text{center}(\mathcal{D}_{i}),
\end{equation}
where $u_{1}, \dots, u_{K}$ denotes the selected representative users, and $D_{i}$ denotes all the historical data that the users have texts on item $i$. We represent the data samples regarding the selected users on item $i$ as: $(u_1, i, y_1), \dots, (u_K, i, y_{K})$.

\subsubsection{Structured Difference Extraction}\label{sec:sde}

After obtaining the representative users, we compare their texts with the target user's texts, \textit{i.e.,} comparing $\{(u_1,i,y_1), \dots, (u_K, i, y_K)\}$ with $(u^\prime, i, y)$ to obtain the difference $d_n$ via an LLM. Instead of allowing LLMs to freely summarize the difference, we would define a standard to structure the LLM difference summarization. The process can be formulated as:
\begin{equation}\label{eq:dif}
\small
\begin{split}
    d_n = \text{LLM}_{\text{dif}}( \{(u_1,i,y_1), \dots, (u_K, i, y_K)\}; (u^\prime, i, y); S),
\end{split}
\end{equation}
% \vspace{-0.5em}
where $\text{LLM}_{\text{dif}}$ denotes the LLM to extract differences.
% and the detailed prompt template is provided in Appendix~\ref{apd_prompt}. 
Here, $S$ represents the standard that the LLMs follow to generate the difference. Specifically, we distill three key dimensions from existing literature (as examples): writing style, emotional style, and semantic style, and then structure the LLM to extract differences from these dimensions. The three dimensions are detailed below:
\vspace{0.6em}
\par
\noindent
\textbf{\textit{Writing style}}. This dimension captures the lexical choices, syntactic structures, and linguistic patterns unique to each user. By analyzing \textit{vocabulary richness, sentence complexity, and grammatical preferences}, we can identify how a user's writing diverges from others \cite{writing}.

\vspace{0.6em}
\par
\noindent
\textbf{\textit{Emotional style}}. This feature reflects the sentiment and affective tone embedded in user's text. By assessing the \textit{polarity (positive, negative, neutral)}, we can highlight differences in how users convey their attitudes and feelings \cite{emotional}.

\vspace{0.6em}
\par
\noindent
\textbf{\textit{Semantic style}}. This aspect focuses on the depth, clarity, and coherence of meaning in user's text. By evaluating \textit{information density, logical flow, and contextual relevance}, we can differentiate how users structure and convey their ideas \cite{semantic}.

\vspace{0.6em}
\par
Extracting differences along these dimensions ensures meaningful and well-structured results for distinguishing task-relevant content. We guide the LLM to adhere to these dimensions for difference extraction using prompts. All prompts used in DPL are provided in Appendix \ref{apd_prompt}.
\section{Experiments}
In this section, we conduct experiments to answer the following research questions:

\noindent \textbf{RQ1}: How does DPL perform on the personalized text generation task with real-world datasets, in comparison to other baseline methods?

\noindent \textbf{RQ2}: What is the impact of the individual components of DPL on its effectiveness?

\noindent \textbf{RQ3}: How do the specific hyper-parameters of DPL influence its performance?

\noindent \textbf{RQ4}: How does DPL perform across different levels of user uniqueness?

\noindent \textbf{RQ5}: How does DPL qualitatively demonstrate its personalized generation capabilities through concrete case analysis?

\subsection{Experimental Setup}

% \par
% \textbf{Datasets.} 
% In this paper, we select data samples from three categories of the open-source Amazon dataset\footnote{\href{https://amazon-reviews-2023.github.io/}{https://amazon-reviews-2023.github.io/}}, including \textit{Books}, \textit{Movies \& TV}, and \textit{CDs \& Vinyl}. After data cleaning, we construct the \textbf{PersoReview} benchmark. Each sample in the dataset represents a user and includes the meta-information of the item for which the review is to be generated, the user's rating and review title, their profile with past reviews, and the actual review as the ground truth. The overall structure of the \textbf{PersoReview} benchmark follows the LongLaMP benchmark, but we make several modifications to suit our experimental settings. More details about the benchmark can be found in Appendix \ref{apd_dataset}.

\par
\textbf{Datasets.} 
We focus on review generation, a representative personalized text generation task. So we conduct experiments on the Amazon Reviews 2023 dataset~\cite{amazon}\footnote{\url{https://amazon-reviews-2023.github.io/}}, focusing on the categories of \textbf{Movies \& TV}, \textbf{CDs \& Vinyl}, and \textbf{Books}. This dataset aggregates user-item interactions from Amazon, encompassing 
user reviews (ratings, texts, etc.) and item metadata (titles, descriptions, etc.). We preprocess the data into a format suitable for personalized review generation tasks, with details provided in Appendix \ref{apd_dataset}.
Our processed datasets are publicly available on Huggingface\footnote{\url{https://huggingface.co/datasets/SnowCharmQ/DPL-main} \& \url{https://huggingface.co/datasets/SnowCharmQ/DPL-meta}}.

\vspace{0.5em}
\par
\noindent
\textbf{Baselines.}
% We compare our proposed DPL with several existing training-free LLM personalization baseline methods for personalized review text generation task.
% To assess the effectiveness of DPL, we compare it with the following LLM-based methods, providing a detailed comparison in Appendix~\ref{apd_base}.
% \qyl{To assess the effectiveness of DPL, we compare it with the following LLM-based methods, providing more details in Appendix~\ref{apd_base}. To ensure fair comparisons, all methods are implemented using the same template and input format as detailed in Appendix \ref{apd_prompt}. 
% Baseline methods are implemented as described in their original papers, with adaptations for our task.}
We compare DPL with the following methods, with more details provided in Appendix~\ref{apd_base}. Note that all methods are implemented using the same template and input format, as outlined in Appendix~\ref{apd_prompt}, to ensure fair comparisons.

% \begin{itemize}
%     \item \textbf{Non-Perso:} In this baseline, the model is provided only with the target item and the review's metadata (\textit{i.e.}, the title and rating) to construct the input, excluding any user profile data.
%     \item \textbf{RAG}~\cite{longlamp}: The RAG method improves personalization by retrieving the top N reviews of items related to the current item, based on the user's profile data. These reviews are then used to construct the model's input. 
%     \item \textbf{Summ}~\cite{summ}: This baseline summarizes the reviews retrieved from the user's profile using an LLM to generate a user's profile summary. The summary, along with the retrieved reviews, is then used to form the model's input. 
%     \item \textbf{Rec-Summ}~\cite{recsumm}: Due to LLM input length limitations, the Summ method cannot process a user's entire profile at once. Rec-Summ overcomes this by dividing the profile data into segments, recurrently refining the summary with each block until the final version is generated. This final summary is then used just like in the Summ method.
%     % \item \textbf{CICL}~\cite{cicl}: The CICL (Contrastive In-Context Learning) method enhances performance by leveraging contrastive examples in the input context. We apply this approach to personalized text generation tasks by randomly selecting another user who has reviewed the same item and include both users' reviews in the prompt for summary generation.
% \end{itemize}

\begin{itemize}
    \setlength{\itemsep}{0pt}

    \item \textbf{Non-Perso}. 
    This refers to the non-personalized approach, which completely excludes any user-specific information from the model input.
    \item \textbf{RAG}~\cite{longlamp}.
    This is a method that employs a retrieval approach to extract key user-specific information, which is then used as instructional context for personalization.
    \item \textbf{IntSum}~\cite{summ}. 
    This method condenses the user’s retrieved history through summarization and incorporates it into the prompt. We achieve this by employing an LLM to generate a summary of the retrieved history.
    % Building on the retrieval approach, this method further condenses the user’s retrieved history through summarization to distill preferences for LLM use.
    \item \textbf{LLM-TRSR}~\cite{recsumm}.
    % This method improves the Summ method by segmenting the user's history into blocks and introduces two distinct preference summarization paradigms: hierarchical summarization and recurrent summarization. We opt for the recurrent summarization technique due to its superior performance.
    This method improves the IntSum method by segmenting the user's history into blocks and employing a recurrent summarization paradigm to iteratively refine the summary within each new block.
    \item \textbf{CICL}~\cite{cicl1}.
    % This is a Contrastive In-Context Learning method, which leverages contrastive examples to enhance the performance of LLMs. We apply it by randomly selecting a different user who has reviewed the same item, incorporating both users' reviews into the prompt for summary generation.
    % This is a contrastive in-context learning method that utilizes contrastive examples to enhance the performance of LLMs. We apply it to the summarizer by incorporating the reviews of a relevant user into the prompt, thereby improving the quality of summary generation.
    This Contrastive In-Context Learning method is not originally designed for LLM personalization. We adapt it to our experimental scenario by applying it to incorporate the review from another user into the prompt to an LLM summarizer, thereby improving the quality of summary generation.
    \item \textbf{Persona-DB}~\cite{personadb}.
    % This method introduces the collaborative refinement approach, which enable a user to retrieve and integrate information from relevant users. We apply it by selecting a relevant user, retrieving their historical reviews, and incorporating both users' reviews into the prompt for review generation.   
    This method introduces a collaborative refinement approach that allows a user to retrieve and integrate information from relevant peers. We apply it by incorporating the review of a relevant user into the prompt for the LLM generator.
\end{itemize}

\vspace{0.5em}
\par
\noindent
\textbf{Evaluation Metrics.} 
We evaluate the experimental results using two complementary approaches, with additional details provided in Appendix \ref{apd_metric}.

\begin{itemize}
\setlength{\itemsep}{0pt}
% \vspace{-0.2em}
\item \textbf{Conventional Evaluation}. 
Following previous work on personalized text generation~\cite{longlamp}, we employ widely-adopted \textbf{ROUGE-1} \cite{rouge}, \textbf{ROUGE-L}, \textbf{METEOR} \cite{meteor}, and \textbf{BLEU} \cite{bleu} to measure lexical overlap between the generated reviews and ground-truth reviews, offering surface-level comparisons through n-gram matching and semantic alignment mechanisms \cite{aupel}.

\item \textbf{LLM-based Evaluation}.
Following the previous evaluation framework~\cite{restpg}, we employ LLMs to assess the nuanced aspects of personalization, going beyond simple lexical matching. Specifically, the LLM evaluators process four key components -- the generated review, the ground-truth review, review metadata, and item attributes -- and produce raw scores ranging from 0 to 10. These scores are then normalized through min-max scaling to the [0, 1] range, with the evaluation metrics denoted as \textbf{S-72B} and \textbf{S-GPT}, corresponding to the use of \textit{Qwen-2.5-72B-Instruct-AWQ}~\cite{qwen25} and \textit{GPT-4o-mini}~\cite{gpt4}, respectively. 
\end{itemize}

% \vspace{1em}
% \par
% \noindent
% \textbf{Implementation Details.} All baseline methods and our proposed DPL are implemented using the same template and prompt structure, as detailed in Appendix \ref{apd_prompt}. We employ the \textit{Qwen2.5-14B-Instruct} model as the backbone LLM for generating personalized reviews and BM25 \cite{bm25} as the retriever. Additionally, we utilize the \textit{gte-Qwen2-1.5B-instruct} \cite{gteqwen} embedding model to map user reviews into vector space for cluster-based user selection. We select a total of four users for comparison, choosing the one whose embedding is closest to the cluster center from each of the four clusters. More implementation details are included in Appendix \ref{apd_env} and \ref{apd_param}.

\newcommand{\meanstd}[2]{$\scalebox{0.76}{#1}_{\pm \scalebox{0.52}{#2}}$}

\begin{table*}[ht]
    \centering
    \fontsize{8}{9.5}\selectfont
    \caption{Performance comparison between the baselines and our DPL on the three datasets, where the best results are highlighted in bold and sub-optimal results are underlined. 
    % ``R-1'', ``R-L'', ``MET.'', and ``BL.'' are respectively abbreviated as ROUGE-1, ROUGE-L, METEOR, and BLEU for simplicity. 
    Higher values indicate better results for all metrics. The symbol * indicates p-value < 0.05 in t-tests.}
    \renewcommand{\arraystretch}{1.4}
    \setlength{\tabcolsep}{4pt}
    \resizebox{0.99\textwidth}{!}{
    \begin{tabularx}{\textwidth}{>{\raggedright\arraybackslash}XX>{\centering\arraybackslash}ccccccccccc}
        \toprule
        \multicolumn{2}{c}{\textbf{Datasets ($\rightarrow$)}} & \multicolumn{3}{c}{\textbf{Movies \& TV}} & \multicolumn{3}{c}{\textbf{CDs \& Vinyl}} & \multicolumn{3}{c}{\textbf{Books}} \\
        \cmidrule(lr){3-5}
        \cmidrule(lr){6-8}
        \cmidrule(lr){9-11}
        \multicolumn{2}{c}{\textbf{Methods ($\downarrow$)}} & ROUGE-1 & METEOR & S-72B & ROUGE-1 & METEOR & S-72B & ROUGE-1 & METEOR & S-72B \\
        \midrule
        \multicolumn{2}{c}{\textbf{Non-Perso}} & \meanstd{0.2279}{0.0022} & \meanstd{0.1301}{0.0008} & \meanstd{0.5372}{0.0043} & \meanstd{0.2396}{0.0029} & \meanstd{0.1327}{0.0007} & \meanstd{0.5351}{0.0048} & \meanstd{0.2711}{0.0058} & \meanstd{0.1511}{0.0023} & \meanstd{0.5497}{0.0050} \\
        \midrule
        \multicolumn{2}{c}{\textbf{RAG}} & \meanstd{0.2844}{0.0017} & \meanstd{0.1870}{0.0050} & \meanstd{0.5852}{0.0042} & \meanstd{0.2914}{0.0014} & \meanstd{0.1857}{0.0076} & \meanstd{0.5804}{0.0057} & \meanstd{0.3238}{0.0031} & \meanstd{0.2171}{0.0041} & \meanstd{0.6158}{0.0043} \\
        \multicolumn{2}{c}{\textbf{IntSum}} & \meanstd{0.2843}{0.0027} & \meanstd{\underline{0.1949}}{0.0031} & \meanstd{0.6054}{0.0029} & \meanstd{0.2958}{0.0015} & \meanstd{\underline{0.1956}}{0.0031} & \meanstd{0.5986}{0.0032} & \meanstd{0.3183}{0.0033} & \meanstd{0.2280}{0.0043} & \meanstd{0.6284}{0.0040} \\
        \multicolumn{2}{c}{\textbf{LLM-TRSR}} & \meanstd{0.2826}{0.0021} & \meanstd{0.1943}{0.0026} & \meanstd{\underline{0.6097}}{0.0032} & \meanstd{0.2937}{0.0023} & \meanstd{0.1952}{0.0029} & \meanstd{\underline{0.6003}}{0.0037} & \meanstd{0.3140}{0.0034} & \meanstd{0.2250}{0.0034} & \meanstd{\underline{0.6292}}{0.0071} \\
        \multicolumn{2}{c}{\textbf{CICL}} & \meanstd{\underline{0.2874}}{0.0029} & \meanstd{0.1946}{0.0037} & \meanstd{0.6048}{0.0039} & \meanstd{\underline{0.2986}}{0.0020} & \meanstd{0.1946}{0.0048} & \meanstd{0.5967}{0.0052} & \meanstd{0.3262}{0.0046} & \meanstd{0.2339}{0.0033} & \meanstd{0.6250}{0.0080} \\
        \multicolumn{2}{c}{\textbf{Persona-DB}} & \meanstd{0.2833}{0.0028} & \meanstd{0.1930}{0.0044} & \meanstd{0.6068}{0.0044} & \meanstd{0.2935}{0.0013} & \meanstd{0.1912}{0.0045} & \meanstd{0.5974}{0.0062} & \meanstd{\underline{0.3271}}{0.0040} & \meanstd{\underline{0.2368}}{0.0048} & \meanstd{\underline{0.6292}}{0.0062} \\
        \midrule
        \multicolumn{2}{c}{\textbf{DPL (ours)}} & \meanstd{\textbf{0.2940*}}{0.0022} & \meanstd{\textbf{0.1990*}}{0.0033} & \meanstd{\textbf{0.6125*}}{0.0048} & \meanstd{\textbf{0.3072*}}{0.0024} & \meanstd{\textbf{0.2036*}}{0.0035} & \meanstd{\textbf{0.6098*}}{0.0064} & \meanstd{\textbf{0.3318*}}{0.0047} & \meanstd{\textbf{0.2411}}{0.0024} & \meanstd{\textbf{0.6300}}{0.0083} \\
        \bottomrule
        \\
        \toprule
        \multicolumn{2}{c}{\textbf{Datasets ($\rightarrow$)}} & \multicolumn{3}{c}{\textbf{Movies \& TV}} & \multicolumn{3}{c}{\textbf{CDs \& Vinyl}} & \multicolumn{3}{c}{\textbf{Books}} \\
        \cmidrule(lr){3-5}
        \cmidrule(lr){6-8}
        \cmidrule(lr){9-11}
        \multicolumn{2}{c}{\textbf{Methods ($\downarrow$)}}  & ROUGE-L & BLEU & S-GPT & ROUGE-L & BLEU & S-GPT & ROUGE-L & BLEU & S-GPT \\
        \midrule
        \multicolumn{2}{c}{\textbf{Non-Perso}} & \meanstd{0.1269}{0.0022} & \meanstd{0.3463}{0.0098} & \meanstd{0.4132}{0.0046} & \meanstd{0.1273}{0.0027} & \meanstd{0.4783}{0.0170} & \meanstd{0.4215}{0.0011} & \meanstd{0.1457}{0.0034} & \meanstd{1.2616}{0.0349} & \meanstd{0.4642}{0.0033} \\
        \multicolumn{2}{c}{\textbf{RAG}} & \meanstd{0.1431}{0.0043} & \meanstd{1.7188}{0.0758} & \meanstd{0.4336}{0.0137} & \meanstd{0.1418}{0.0032} & \meanstd{1.9271}{0.1491} & \meanstd{0.4456}{0.0079} & \meanstd{0.1636}{0.0034} & \meanstd{4.1082}{0.1215} & \meanstd{0.5077}{0.0066} \\
        \multicolumn{2}{c}{\textbf{IntSum}} & \meanstd{0.1424}{0.0036} & \meanstd{2.0619}{0.0355} & \meanstd{0.4345}{0.0093} & \meanstd{0.1422}{0.0031} & \meanstd{2.3621}{0.0775} & \meanstd{\underline{0.4482}}{0.0034} & \meanstd{0.1614}{0.0030} & \meanstd{4.6256}{0.2021} & \meanstd{0.5102}{0.0044} \\
        \multicolumn{2}{c}{\textbf{LLM-TRSR}} & \meanstd{\underline{0.1439}}{0.0039} & \meanstd{\underline{2.0834}}{0.0298} & \meanstd{0.4350}{0.0077} & \meanstd{0.1417}{0.0032} & \meanstd{\underline{2.3628}}{0.0538} & \meanstd{0.4472}{0.0034} & \meanstd{0.1603}{0.0034} & \meanstd{4.6268}{0.2724} & \meanstd{0.5019}{0.0056} \\
        \multicolumn{2}{c}{\textbf{CICL}} & \meanstd{0.1439}{0.0039} & \meanstd{2.0576}{0.0795} & \meanstd{0.4353}{0.0094} & \meanstd{\underline{0.1433}}{0.0033} & \meanstd{2.2978}{0.1200} & \meanstd{0.4454}{0.0062} & \meanstd{0.1664}{0.0041} & \meanstd{5.1555}{0.2748} & \meanstd{0.5090}{0.0048} \\
        \multicolumn{2}{c}{\textbf{Persona-DB}} & \meanstd{0.1425}{0.0037} & \meanstd{1.9898}{0.0777} & \meanstd{\underline{0.4369}}{0.0082} & \meanstd{0.1421}{0.0033} & \meanstd{2.1616}{0.1455} & \meanstd{0.4467}{0.0069} & \meanstd{\underline{0.1672}}{0.0036} & \meanstd{\underline{5.1862}}{0.2064} & \meanstd{\underline{0.5123}}{0.0085} \\
        \multicolumn{2}{c}{\textbf{DPL}} & \meanstd{\textbf{0.1466*}}{0.0035} & \meanstd{\textbf{2.2981*}}{0.0639} & \meanstd{\textbf{0.4437*}}{0.0057} & \meanstd{\textbf{0.1463*}}{0.0037} & \meanstd{\textbf{2.6803*}}{0.0559} & \meanstd{\textbf{0.4556*}}{0.0052} & \meanstd{\textbf{0.1720*}}{0.0045} & \meanstd{\textbf{5.9379*}}{0.3108} & \meanstd{\textbf{0.5224*}}{0.0091} \\
        \bottomrule
        
    \end{tabularx}
    }
    \label{main_table}
\end{table*}

\vspace{0.5em}
\par
\noindent
\textbf{Implementation Details}.
% To ensure fair comparisons, all methods are implemented using the same template and prompt structure as detailed in Appendix \ref{apd_prompt}. For the baseline methods, we implement them according to the respective descriptions in their original papers.
For our DPL, we use the \textit{Qwen2.5-14B-Instruct}~\cite{qwen25} model as the LLM backbone, which functions as the difference extractor, summarizer, and review generator. 
Following the setting of LongLaMP~\cite{longlamp}, we retrieve user reviews as auxiliary signals using BM25, where the retrieval query is formed by concatenating a short review title with the item's description.
For selective user comparison and structured difference extraction, we set the number of retrieved reviews, \textit{i.e.}, $N$ in Equation~\eqref{eq:sum}, to 8.
Additionally, we utilize \textit{gte-Qwen2-1.5B-Instruct}~\cite{gteqwen} to map user reviews into embeddings, which are subsequently clustered into 5 groups by default, equivalent to setting the number of selected representative users $K$ in Equation~\eqref{eq:cluster} to 4. 
To assess the reliability of our generated results, we conduct inference using different temperature values (0.2, 0.4, 0.6, 0.8, and 1.0) for each method, and report the mean and standard deviation of the evaluation metrics. We also perform t-tests to assess statistical significance.
More details are provided in Appendix \ref{apd_imp}.

\subsection{Main Results (RQ1)}

% In this subsection, we study the performance of our proposed DPL method across three categories of the PersoReview benchmark. For both the personalized baseline methods and our approach, we consider four scenarios based on the number of historical review data points retrieved from the user's profile: 1, 2, 4, and 8. The main experimental results are shown in Table \ref{main_table}. From the table, we have the following observations:

We begin by evaluating the overall performance of the compared methods. The summarized results are presented in Table~\ref{main_table}, from which the following observations can be drawn:

% \vspace{-0.4em}
\begin{itemize}
\setlength{\itemsep}{0pt}

\item 
DPL demonstrates consistently superior personalization performance across all three datasets and all evaluation metrics.
For instance, on the CDs \& Vinyl category, DPL achieves the highest scores on all metrics, with an average relative improvement of 4.29\% over the best baselines and six metrics showing statistically significant gains.
Similar trends are observed on Movies \& TV and Books, where DPL also ranks first across all metrics, with 6 and 4 metrics achieving significance, respectively.
This result underscores its enhanced personalized text generation capabilities, which can be attributed to its focus on extracting inter-user differences to improve LLM personalization through the selective comparison mechanism and difference extraction standard.
% DPL demonstrates superior personalization performance compared to the baseline methods across all metrics on three datasets. For instance, on the CDs \& Vinyl dataset, it shows an average improvement of 2.85\% over the best baseline metrics. This result underscores its enhanced personalized text generation capabilities, which can be attributed to its focus on extracting inter-user differences to improve LLM personalization through the selective comparison mechanism and difference extraction standard.

\item 
Non-Perso exhibits the weakest personalization performance across all experimental settings, which can be attributed to its fundamental limitation in incorporating user-specific contextual signals. In comparison, RAG demonstrates substantial performance gains by leveraging a retrieval approach to integrate user-specific information, thereby enabling more tailored outputs.

\item 
IntSum and LLM-TRSR demonstrate consistent superiority over RAG across all datasets, primarily due to their integration of summarization mechanisms. This is because summarization mechanisms distill the retrieved user history into more concise representations, thereby enhancing the utilization of contextual information.

\item 
Although CICL and Persona-DB demonstrate some performance improvements by utilizing the historical reviews of other relevant users, they still lag behind our DPL. This discrepancy arises because they fail to incorporate inter-user comparative analysis, which is essential for identifying the nuanced differences that genuinely influence user preferences.

\end{itemize}

\subsection{Ablation Studies (RQ2)}

% In this subsection, we conduct additional ablation studies to further investigate the effects of various DPL design choices, including the number of users selected for comparison, the user selection method, and the influence of stylistic features. To reduce resource consumption, our analysis primarily focuses on the \textit{Books} dataset and uses 8 retrieved historical review data points.

To substantiate the rationale behind the various design decisions of DPL, we conduct an exhaustive evaluation using the Books dataset by systematically disabling one critical design element to obtain various variants.
For the \textbf{selective user comparison} in Section~\ref{sec:suc}, we replace the clustering-based selection method and introduce the following two variants for comparison:

\vspace{-0.5em}
\begin{itemize}
\setlength{\itemsep}{0pt}

\item \textbf{Random}. 
This variant randomly selects users who have reviews on the same retrieved historical item without considering any specific criteria.

\item \textbf{SimRank}. 
This variant selects users based on the similarity of historical reviews between the current user and potential candidates. Specifically, the users with the top-$K$ lowest similarity are chosen in order to maximize differentiation.

\end{itemize}

\vspace{-0.5em}
For the \textbf{structured difference extraction} in Section~\ref{sec:sde}, we modify the original extraction standard, whose dimensions include writing style, emotional style, and semantic style, and introduce the following four variants for comparison:

\vspace{-0.5em}
\begin{itemize}
\setlength{\itemsep}{0pt}

\item \textbf{None}. 
This variant excludes all the dimensions.

\item \textbf{WriOnly, EmoOnly, SemOnly}. 
These three variants each incorporate one of the three style dimensions, respectively.

\end{itemize}

\begin{table}[t]
    \caption{Results of the ablation study for DPL on Books.}
    \label{ablation-study}
    \centering
    \small
    \resizebox{0.48\textwidth}{!}{
    \begin{tabular}{cccc}
        \toprule
        \textbf{Methods} & R-1  & MET. & S-72B \\
        \midrule
        \textbf{DPL} & \meanstdllm{\textbf{0.3318}}{0.0047} & \meanstdllm{\underline{0.2411}}{0.0024} & \meanstdllm{\textbf{0.6300}}{0.0083} \\
        \midrule
        \textbf{Random} & \meanstdllm{0.3299}{0.0034} & \meanstdllm{0.2379}{0.0049} & \meanstdllm{\underline{0.6269}}{0.0063} \\
        \textbf{SimRank} & \meanstdllm{0.3299}{0.0037} & \meanstdllm{0.2400}{0.0059} & \meanstdllm{0.6235}{0.0031} \\
        \midrule
        \textbf{None} & \meanstdllm{0.3235}{0.0034} & \meanstdllm{0.2335}{0.0056} & \meanstdllm{0.6149}{0.0028} \\
        \textbf{WriOnly} & \meanstdllm{\underline{0.3303}}{0.0043} & \meanstdllm{\textbf{0.2427}}{0.0048} & \meanstdllm{0.6188}{0.0028} \\
        \textbf{EmoOnly} & \meanstdllm{0.3238}{0.0045} & \meanstdllm{0.2334}{0.0028} & \meanstdllm{0.6262}{0.0034} \\
        \textbf{SemOnly} & \meanstdllm{0.3246}{0.0040} & \meanstdllm{0.2387}{0.0052} & \meanstdllm{0.6260}{0.0035} \\
        \bottomrule
    \end{tabular}}
\end{table}

\vspace{-0.5em}
Table~\ref{ablation-study} illustrates the comparison results on ROUGE-1, METEOR, and S-72B, from which we draw the following observations:

\vspace{-0.5em}
\begin{itemize}
\setlength{\itemsep}{0pt}

\item 
For the selective user comparison, replacing the clustering-based selection method results in a performance decline. This can be attributed to the fact that random user selection does not rely on any criteria, resulting in insufficient comparative information. Additionally, although SimRank introduces pairwise similarity for enhanced user selection, it fails to capture the global context offered by clustering, leading to a lack of representativeness in the selected users.

\item 
For the structured difference extraction, removing parts of the extracted dimensions leads to a general performance drop. This finding suggests that each dimension plays a crucial role in capturing different facets of the differences, thereby supporting meaningful and well-structured extraction for distinguishing task-relevant content.
\end{itemize}

\begin{figure}[t]
    \centering
    % \includesvg[width=1\linewidth]{figs/ablation-user-num.svg}
    \includegraphics[width=0.96\linewidth]{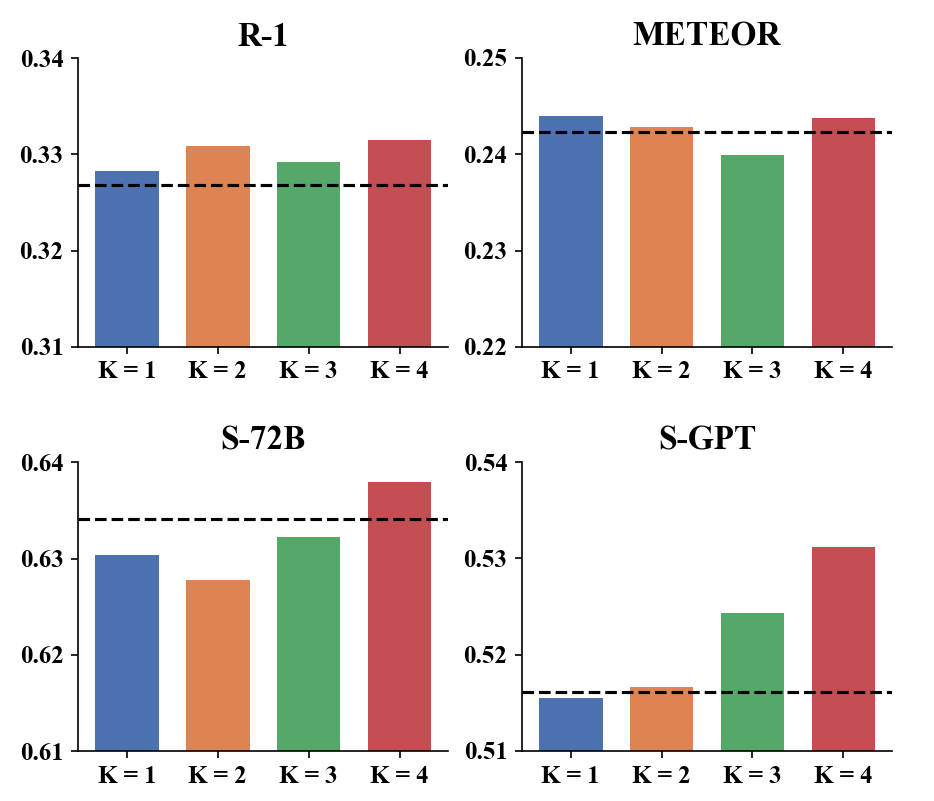}
    \caption{Results of the performance of DPL across
    different values of $K$, where $K$ denotes the number of selected representative users. The dashed line represents the corresponding best baseline metric.}
    \label{ablation-user-num}
\end{figure}
\begin{figure}[t]
    \centering
    \includegraphics[width=0.96\linewidth]{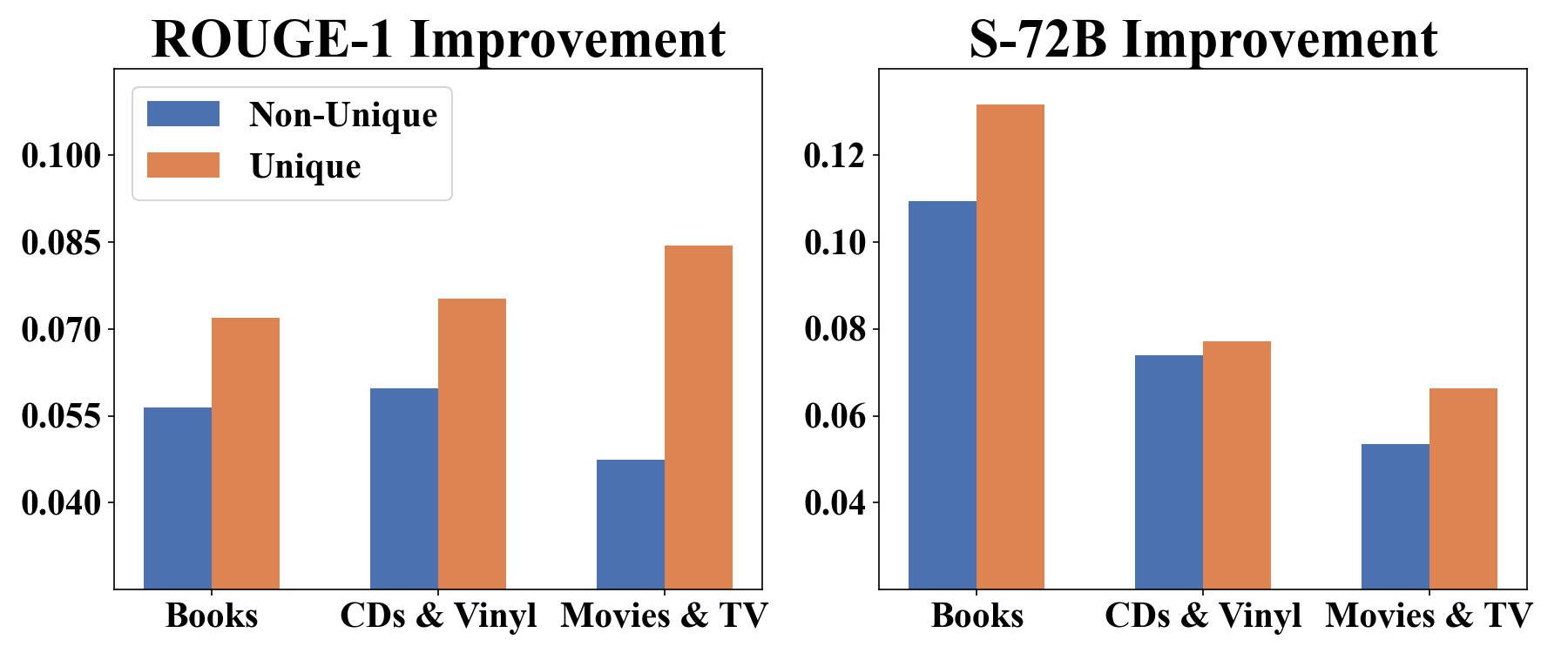}
    \caption{Results of the performance of DPL across different levels of uniqueness. We report the absolute improvement compared to the non-personalization method.}
    \label{in-depth}
    % \vspace{-1em}
\end{figure}

\subsection{In-depth Analyses (RQ3 \& RQ4)}
We next evaluate the performance of DPL across different values of hyperparameters and different levels of user uniqueness. 
\subsubsection{Hyper-parameter Analysis}

% In our investigation, the two hyperparameters, $N$ and $K$, play pivotal roles in determining the effectiveness of DPL. Specifically, $N$ represents the number of retrieved key elements from the user history, while $K$ denotes the number of selected central users for comparison. We conduct a systematic examination to assess the impact of varying these parameters on performance. Additionally, we present the performance of the best baseline for comparative analysis. In particular, we vary $N$ within the set $\{1, 2, 4, 8\}$ and $K$ within the set $\{1, 2, 3, 4\}$. Here, we present the results for varying $K$ in Figure~\ref{ablation-user-num} and defer the results for varying $N$ to Appendix~\ref{}.

In our investigation, the hyperparameter $K$, which denotes the number of selected representative users for comparison, plays a pivotal role in determining the effectiveness of DPL. We vary $K$ within the set $\{1, 2, 3, 4\}$ to assess the impact of varying these parameters on performance. Additionally, we present the best baseline metrics for comparative analysis, with the results shown in Figure~\ref{ablation-user-num}, using a fixed temperature of 0.8. Our findings indicate that, generally, increasing the number of representative users leads to slight to moderate improvements in performance, with the best results occurring at $K=4$. This underscores that effective difference extraction depends on selecting an appropriate number of comparative representatives.

% We observe that DPL consistently demonstrates superior performance compared to the baseline across all values of $K$, validating its robustness in personalized review generation. 
% Furthermore, the trends across different metrics are not uniform, which may be attributed to the distinct aspects of generation qualities that these metrics capture.

\subsubsection{Impact of User Uniqueness}

We initially generate the uniqueness label based on users' historical review data, as it does not inherently exist. Specifically, we map each user's historical reviews to embeddings using LLMs and compute the average embedding of all users to establish a center. The Euclidean distance between each user's embedding and this center is then calculated. Users with distances in the bottom 50\% are classified as \textbf{non-unique}, while those in the top 50\% are classified as \textbf{unique}. 

Following this classification, we compare the performance of DPL with the non-personalization method across the two groups. For simplicity, we report the absolute improvement of ROUGE-1 and S-72B, as the trends for other metrics are consistent. As illustrated in Figure~\ref{in-depth}, the unique group shows a significantly greater improvement than the non-unique group. This suggests that the more unique the user, the better we can capture insights into their differences relative to others, aligning with the philosophy that ``differences make us unique''.

\subsection{Case Study (RQ5)}
\begin{figure}[t]
    \centering
    \includegraphics[width=0.91\linewidth]{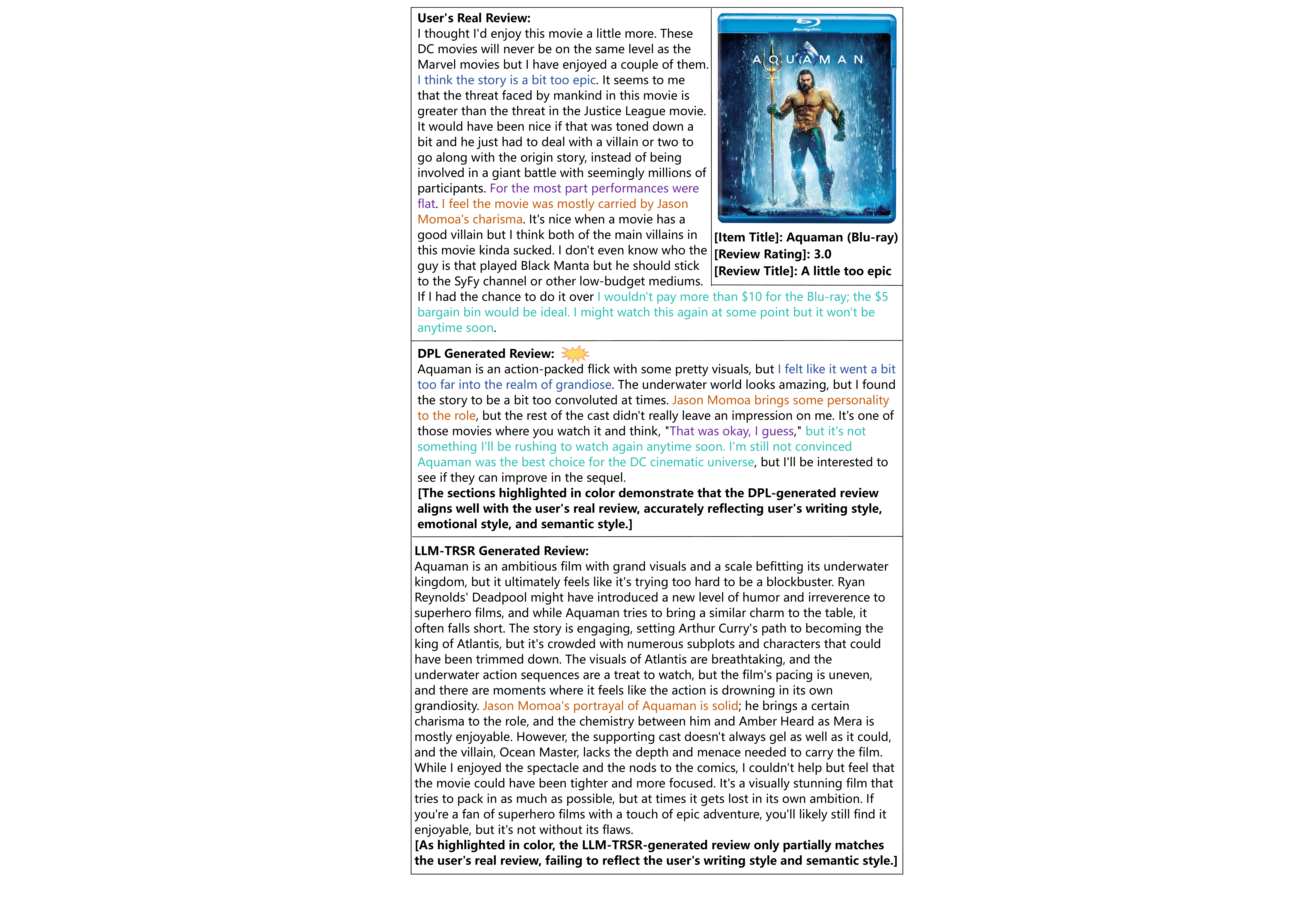}
    % \vspace{-5pt}
    \caption{Comparison of item reviews generated by DPL and the baseline method LLM-TRSR, along with the user's real review. The consistent sections between the generated reviews and the user's real review are highlighted by specific colors.}
    \label{case-study-main}
    \vspace{-1em}
\end{figure}

Finally, we conduct a case study to gain a deeper understanding of DPL's capacity for personalized review generation. Specifically, we select a representative item from the Movies \& TV dataset and compare a single review generated by DPL with that produced by the representative baseline method, LLM-TRSR. The detailed reviews are presented in Figure~\ref{case-study-main}, where we highlight consistent sections between the generated reviews and the user’s real review by distinct colors. 

We observe that the review generated by DPL exhibits more consistent sections. Furthermore, it offers a more personalized perspective and concise content, whereas the review generated by LLM-TRSR adopts a more objective stance and contains more tedious information. The alignment of DPL with the user's writing style and semantic preferences further supports its superiority. More case studies are provided in Appendix \ref{apd_case}.

\section{Related Work}

With the widespread use of LLMs~\cite{zhao2024comprehensive,binllm,sipo,zhang2025reinforced,sarft,lmtp,dependeval,alphaedit}, there is a growing need for personalized responses tailored to individual users.
LLM personalization has been explored across various domains, underscoring its importance~\cite{cso,prefeval,personajudge,puma}. 
% PUMA~\cite{puma} introduces PersonalWAB for evaluating personalized web tasks, while PMG~\cite{pmg} extends personalization to multimodal generation. 
While existing works advance personalization in different scenarios, such as recommendation~\cite{sdpo,d3,agent4rec} and image generation~\cite{pmg,pigeon,drc}, personalized text generation remains the foundation of LLM personalization~\cite{zhao2025nextquill,persoSurvey1}.
LaMP~\cite{lamp} evaluates short-form personalized text, later extended by LongLaMP~\cite{longlamp} to assess long-form text generation. Among these, personalized item review writing stands out as a key task~\cite{reviewllm,pgraphrag} since user reviews naturally reflect individual preferences~\cite{amazon20181}, serving as a critical scenario to evaluate a model's ability to generate personalized content.

% Although fine-tuning~\cite{perpcs, personpeft} and reinforcement learning from human feedback~\cite{personrlhf} are effective approaches for LLM personalization, they require separate model's parameters for each user, increasing costs. 
% A more efficient alternative is
Recent LLM-based methods enhance personalization using user-specific content, %\qyl{falling into two categories:}
which can be categorized into:
1) retrieval-based methods, where ROPG \cite{ropg} introduces a retrieval selection model to dynamically choose the best retriever and HYDRA \cite{hydra} employs a personalized reranker to prioritize useful information; and 2) summarization-based methods, where ONCE \cite{once} generates user profiles by summarizing topics and regions of interest in their browsing history
% where ONCE \cite{once} generates user profiles by analyzing their browsing history to summarize topics and regions of interest 
and PPlug \cite{pplug} converts user history into personalized embeddings.
% uses an embedder to convert a user's history into a personalized embedding.

% ROPG \cite{ropg} introduces a retrieval selection model to dynamically choose the best retriever. HYDRA \cite{hydra} employs a personalized reranker to prioritize useful information. PPlug~\cite{pplug} uses an embedder to convert a user's history into a personalized embedding, which is added to the task input to improve preference understanding.

% A recent work Persona-DB~\cite{personadb} has explored using data from other users to enhance the personalized text generation for the current user. However, their approach primarily focuses on enriching relevant knowledge from other users rather than truly capturing what makes each user unique.

% \xyz{A recent study \cite{cicl} has explored contrastive learning to incorporate inter-user data by using positive examples to enhance learning and negative examples to enforce distinction. 
% However, in user-centered LLM personalization, the most natural positive example is the user’s data, while identifying other relevant negative examples from unsupervised data remains challenging in the realistic application. This limitation hinders the automatic and comprehensive capture of user-relevant inter-user information. Therefore, effectively utilizing inter-user information remains an open challenge, leaving a critical gap in LLM personalization.}

Besides, a recent study Persona-DB~\cite{personadb} uses data from other users to enhance personalized text generation. However, this method mainly focuses on enriching relevant knowledge rather than capturing each user's uniqueness, key to effective personalization.
To leverage inter-user data, contrastive learning \cite{cicl1} offers a promising way of using both positive and negative examples to better describe intents. 
% However, its performance is contingent on the identification of meaningful negative examples, which are challenging to obtain in user-centered LLM personalization.
However, its performance relies on meaningful negative examples, which are challenging to obtain in user-centered LLM personalization.
This reliance limits the full leverage of inter-user information for LLM personalization. To the best of our ability, our proposed DPL addresses this limitation.

\section{Conclusions}
% In this work, we introduced DPL, a novel method that enhances LLM personalization by identifying and leveraging inter-user differences as key preference information. We proposed a selective user comparison mechanism and a structured difference extraction method to extract task-relevant meaningful differences effectively. Experimental results on the personalized review generation benchmark show that DPL significantly enhances LLM personalization.
% Currently, we only focus on review generation tasks to validate our method. In future, we plan to extend it to other tasks for broader verification. Additionally, our method's effectiveness depends on the long-context processing capabilities of LLMs, and we currently assume that the context window is sufficient for handling our data. In future, we would extend our method to address long-context cases.

This study introduced DPL, a novel method that enhances LLM personalization by identifying and leveraging inter-user differences as key preference information. We proposed a selective user comparison mechanism and a structured difference extraction method to extract task-relevant meaningful differences effectively. Experimental results on the personalized review generation task show that DPL significantly enhances LLM personalization.
Currently, we only focus on review generation tasks to validate our method. In future, we plan to extend it to other tasks for broader verification. Additionally, our method's effectiveness depends on the long-context processing capabilities of LLMs, and we currently assume that the context window is sufficient for handling our data. We also plan to extend our method to address long-context cases.
% \qyl{Currently, we only focus on review generation tasks to validate our method. Meanwhile, our approach relies on the long-context processing capabilities of LLMs, and we assume that the context window is sufficient for handling our data. In future, we plan to expand it to more tasks for broader verification and adapt our method for long-context cases.}

% In this work, we introduced DPL, a novel method that enhances LLM personalization by identifying and leveraging inter-user differences as key preference information.  We proposed a selective comparison mechanism for choosing representative users to compare and a structured difference extraction method to extract task-relevant and meaningful differences effectively. Experimental results on the personalized review generation benchmark demonstrate that DPL significantly enhances LLM personalization. 
% Currently, we focus on review generation tasks to validate our method, but we plan to extend it to other tasks for broader verification. Moreover, our method's effectiveness relies on the long-context processing abilities and we assume that the context window of LLMs is enough to deal with our data. In the future, we would extend our method to deal with long context cases.
% Additionally, our method relies on the LLM's ability to leverage preference context for personalization, which may be ineffective if the LLM lacks sufficient capability. In the future, we will explore directly enhancing inherent personalization abilities without difference-aware preference modeling.
\section*{Limitations}

In this paper, we utilize LLMs to extract inter-user differences, summarize difference-aware representations, and generate personalized reviews. These processes require a highly powerful model to effectively analyze users' personalized information, ensuring that the generated reviews align closely with individual user preferences. Moreover, in this task, we specifically focus on three key aspects: writing style, emotional style, and semantic style. Our findings indicate that these factors significantly enhance the quality of personalized reviews. However, in other tasks, these standards may not be as suitable, and additional factors, such as personality traits, might need to be considered. Our approach involves comparing differences between users, which inevitably introduces additional computational overhead and time costs. We believe this challenge can be mitigated by precomputing and storing difference-aware representations in user memory. These representations can then be directly retrieved during online deployment, with periodic updates ensuring both efficiency and relevance.

% Currently, this paper has certain limitations that need to be addressed in future work. First, our method requires a highly powerful LLM to effectively analyze user's personalized information. Second, the structured standard primarily focuses on three key aspects -- writing style, emotional style, and semantic style -- which may not be as effective in other tasks, where additional elements, such as personality traits, could be more relevant. Last, comparing differences between users introduces additional computational overhead and time costs. We believe this challenge can be mitigated by precomputing and storing difference-aware representations in user memory. These representations can then be directly retrieved during online deployment, with periodic updates ensuring both efficiency and relevance.
\section*{Ethical Statements}

% Our approach to personalization involves storing and retrieving user historical interaction data, requiring strict safety protocols and ethical considerations during deployment. Since we leverage inter-user differences to enhance personalization, sharing and integrating information between users necessitates careful attention to privacy. This process must incorporate robust privacy safeguards, including data anonymization, secure storage, and explicit user consent. Furthermore, when deploying applications for personalization, the ability to create detailed user profiles raises concerns about potential misuse, such as surveillance or targeted manipulation. To avoid these risks, developers have a responsibility to prioritize privacy-preserving techniques and ensure that user data is handled with integrity and respect.

Our method presents the following potential ethical concerns that need to be addressed carefully during deployment. First, it involves the storage and retrieval of users' historical interaction data, which necessitates the implementation of strict safety protocols and ethical guidelines. Second, as we leverage inter-user differences to enhance personalization, sharing and integrating information between users raises significant privacy concerns. This process must incorporate robust privacy safeguards, including data anonymization, secure storage, and explicit user consent. Furthermore, the ability to create detailed user profiles for personalization purposes introduces the risk of potential misuse, such as surveillance or targeted manipulation. To mitigate these risks, developers bear the responsibility to prioritize privacy-preserving techniques and ensure that user data is handled with the utmost integrity and respect.

In our research, we use an open-source dataset, ensuring full compliance with the original dataset's MIT license. Our research prioritizes data integrity, privacy, and ethical considerations to uphold the responsible and fair use of open-source resources.
% \section*{Acknowledgments}

\bibliography{arxiv_latex}
\appendix

\section{Dataset Details}\label{apd_dataset}

In this paper, we focus on item review text generation, which is an effective task to evaluate the model's ability to generate personalized content. To generate data samples, we employ the open-source Amazon Reviews Dataset~\cite{amazon} and primarily follow the setting of LongLaMP~\cite{longlamp}, with several modifications to better align with our experimental framework.

\vspace{0.6em}
\par
\noindent
\textbf{Dataset Introduction.} The Amazon Reviews dataset is processed by merging the user reviews subset with the item metadata subset using the \textit{asin} field as the shared field. Specifically, the \textit{asin} value from the user reviews subset is matched to the corresponding \textit{parent\_asin} field in the item metadata subset. From the user reviews subset, we extract \textit{asin}, which is the unique identifier for the item; \textit{user\_id}, representing the unique identifier for the reviewer; \textit{title}, which denotes the title of the review; \textit{text}, containing the actual review content; \textit{rating}, indicating the numerical rating given by the user; and \textit{timestamp}, specifying the time when the review was submitted. From the item metadata subset, we select \textit{parent\_asin}, which serves as a reference to the \textit{asin} field in the user reviews subset, linking reviews to items; \textit{title}, representing the title of the item; and \textit{description}, providing a textual summary of the item's features. To distinguish between the review title and the item title, we use \textit{review\_title} for the former and \textit{item\_title} for the latter.

\begin{figure}[t]
    \centering
    \includegraphics[width=1\linewidth]{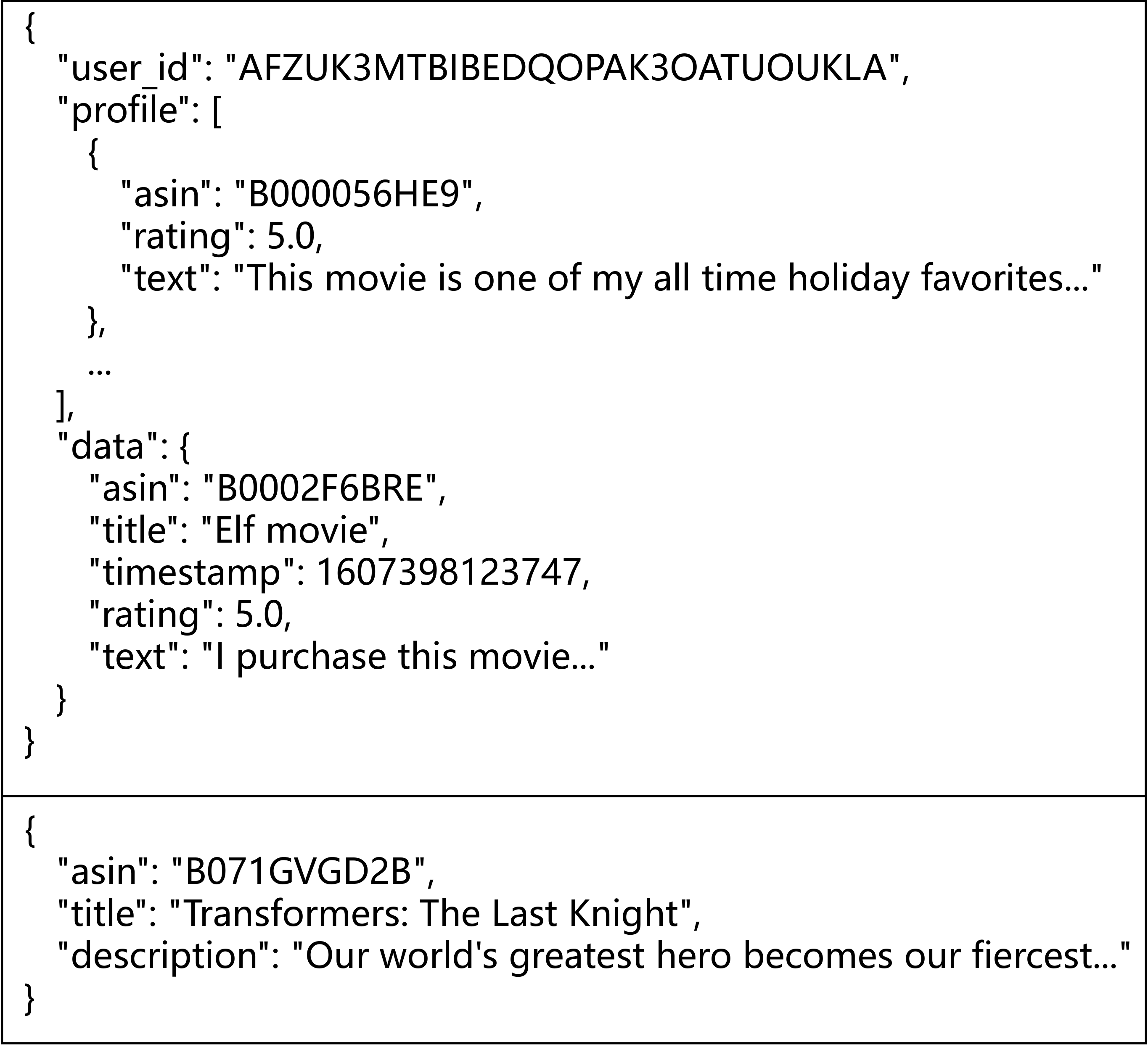}
    \caption{Examples of the main experimental dataset (above) and the item dataset (below).}
    \label{dataset-demo}
\end{figure}

\begin{table}[t]
    \centering
    \caption{Dataset statistics of three categories in our processed Amazon dataset.}
    \renewcommand{\arraystretch}{1.1}
    \setlength{\tabcolsep}{6pt} 
    \resizebox{0.48\textwidth}{!}{
    \begin{tabular}{ccccc}
        \toprule
        \multicolumn{2}{c}{\textbf{Categories ($\downarrow$)}} & \#data & Profile Size & Output Length \\
        \midrule
        \multirow{3}{*}{\textbf{\shortstack{Movies \\ \& TV \\ (4832)}}} & Train & 15400 & 35.61$\pm$35.97 & 1691.42$\pm$1617.71 \\
        & Val & 1925 & 40.11$\pm$35.89 & 1668.26$\pm$1612.17 \\
        & Test & 1925 & 41.11$\pm$35.89 & 1704.61$\pm$1751.98 \\
        \midrule
        \multirow{3}{*}{\textbf{\shortstack{CDs \& \\ Vinyl \\ (3801)}}} & Train & 14032 & 33.00$\pm$32.44 & 1485.19$\pm$1292.09 \\
        & Val & 1754 & 37.50$\pm$32.36 & 1602.93$\pm$1431.09 \\
        & Test & 1754 & 38.50$\pm$32.36 & 1600.04$\pm$1419.49 \\
        \midrule
        \multirow{3}{*}{\textbf{\shortstack{Books\\ (839)}}} & Train & 2536 & 29.34$\pm$22.63 & 1295.77$\pm$827.48 \\
        & Val & 317 & 33.84$\pm$22.51 & 1216.49$\pm$768.44 \\
        & Test & 317 & 34.84$\pm$22.51 & 1194.90$\pm$801.18 \\
        \bottomrule
    \end{tabular}}
    \label{dataset-stat}
\end{table}

\begin{table*}[t]
    \caption{Comparison of different baseline methods.}
    \label{method-comparison}
    \centering
    % \small
    \resizebox{0.95\textwidth}{!}{\begin{tabular}{l|cccc}
        \toprule
        Method     & Training-Free & Current User's Data & Other Users' Data & Difference-Aware \\
        \midrule
        Non-Perso  & \cmark    & \xmark         & \xmark        & \xmark       \\
        RAG        & \cmark    & \cmark         & \xmark        & \xmark       \\
        IntSum       & \cmark    & \cmark         & \xmark        & \xmark       \\
        LLM-TRSR   & \cmark    & \cmark         & \xmark        & \xmark       \\
        CICL       & \cmark    & \cmark         & \cmark        & \xmark       \\
        Persona-DB & \cmark    & \cmark         & \cmark        & \xmark       \\
        DPL        & \cmark    & \cmark         & \cmark        & \cmark       \\
        \bottomrule
    \end{tabular}
    }
\end{table*}

\vspace{0.6em}
\par
\noindent
\textbf{Data Curation.} 
We perform a detailed data-cleaning process to create a high-quality benchmark. First, we ensure that each review includes a corresponding \textit{review\_title}, \textit{rating}, and \textit{timestamp}, with a minimum \textit{text} length of 200 characters to meet the long-form criteria. For each item, we verify that it has an \textit{item\_title} and that its \textit{description} falls within the 100 to 2000-character range to maintain high-quality content. Following this, we ensure that each product has at least four unique reviewers and retain only the most recent review from each reviewer for a given item. Additionally, we restrict the number of reviews per reviewer to between 18 and 500 to filter out users who may engage in review manipulation or have insufficient data for meaningful personalization. 
To meet the requirements of our experimental setup, we further ensure that each user in the validation and test sets has at least 8 historical records, where each record corresponds to an item that has been reviewed by at least 5 users, including the target user. This constraint guarantees that we can retrieve a sufficient number of other users for computing difference-aware representations during evaluation.
These settings are designed to align with our experiments.

\vspace{0.6em}
\par
\noindent
\textbf{Dataset Construction.}
After preprocessing, we proceed to construct the main experimental dataset. Here we follow the Temporal Setting used in LongLaMP, treating each user as a data point. However, unlike LongLaMP, we take a different approach when constructing the training dataset. Specifically, we use a user's 10th most recent to 3rd most recent reviews as training samples, with their profile information derived from historical reviews relative to the target review. The validation dataset consists of the user's 2nd most recent review, while the test dataset includes the 3rd most recent review. Each user's historical information samples include only the item \textit{asin}, \textit{rating}, \textit{timestamp}, \textit{review\_title}, and \textit{text}. The \textit{asin} references the pre-filtered item dataset, allowing us to extract the corresponding item information. Examples of the main experimental dataset and the item dataset are illustrated in Figure \ref{dataset-demo}. We process the three categories: Books, Movies \& TV, and CDs \& Vinyl separately. The statistics for these categories after dataset construction are presented in Table \ref{dataset-stat}.

\vspace{0.6em}
\par
\noindent
\textbf{Dataset Usage.} 
In our experiments, we utilize only the test dataset for a direct evaluation of model performance. The details of model inputs are provided in Appendix \ref{apd_prompt}. Notably, the training and validation datasets are included to support related research.

\section{Baselines Details}\label{apd_base}

In this section, we further introduce each baseline method in detail. Additionally, the comparison between different baseline methods and our proposed DPL is shown in Table \ref{method-comparison}.

\begin{itemize}
\setlength{\itemsep}{0pt}

    \item \textbf{Non-Perso:} This method does not utilize any user-specific information to assist in personalized review generation. The model's input consists only of the item's title and description, along with the rating and title of the review to be generated.
    \item \textbf{RAG:} This method employs a retriever, specifically BM25 in our experiments, to retrieve relevant past reviews from the user's review history. These retrieved reviews are then formatted and incorporated into the LLM's input prompt.
    \item \textbf{IntSum:} Unlike the original implementation of generating a summary offline and directly using it in downstream tasks, our scenario of long-form text review generation also involves long-form user history. Due to the input length limitations of LLMs, we cannot feed all of a user's historical reviews into the model at once to produce a sensible user profile summary. Therefore, we explore an alternative solution: instead of generating a static summary beforehand, the model dynamically creates a summary based on the retrieved user's historical reviews. This summary is then integrated into the model's input prompt along with retrieved historical reviews. The experimental results demonstrate that our implementation is correct.
    \item \textbf{LLM-TRSR:} To address the issue that our IntSum experiment does not utilize all of the user's historical data, we select the LLM-TRSR method as a baseline. This method divides the user's history into blocks and introduces two distinct preference summarization paradigms: hierarchical summarization and recurrent summarization. We opt for the recurrent summarization technique due to its superior performance and better suitability for the temporal setting of our dataset. In this method, a summary is first generated based on the initial history block and then iteratively updated using more recent history blocks. The final summary is then integrated into the model's input along with the retrieved user's historical reviews. In our experiments, each history block contains eight reviews.
    \item \textbf{CICL:} This is the Contrastive In-Context Learning method. While originally not designed for LLM personalization, we adapt it to our experimental scenario. In our experiments, we treat the current user's review as the positive example and a review from another user as the negative example. To construct these comparisons, we use the BM25 retriever to retrieve a historical review from the current user and randomly select a review on the same item from another user. These comparative reviews, along with the current user's past reviews, are incorporated into the prompt for an LLM summarizer, thereby improving the quality of summary generation.
    \item \textbf{Persona-DB:} This method introduces a collaborative refinement approach that allows a user to retrieve information from relevant peers to enrich knowledge. We adapt this approach to our task. For selecting relevant users, we employ an embedding-based strategy using the gte-Qwen2-1.5B-Instruct embedding model to encode users' historical information and identify the most similar user to the current user. When generating the review, we first create a summary based on the current user's retrieved historical reviews. Simultaneously, we extract the same number of reviews from the most similar user's history. The current user's profile summary, historical reviews, and the most similar user's historical reviews are then integrated into the prompt for the LLM generator.
    
\end{itemize}

All baseline methods, including our approach, use prompts that are uniformly designed for our review generation task.

\begin{figure}[t]
    \centering
    \includegraphics[width=1\linewidth]{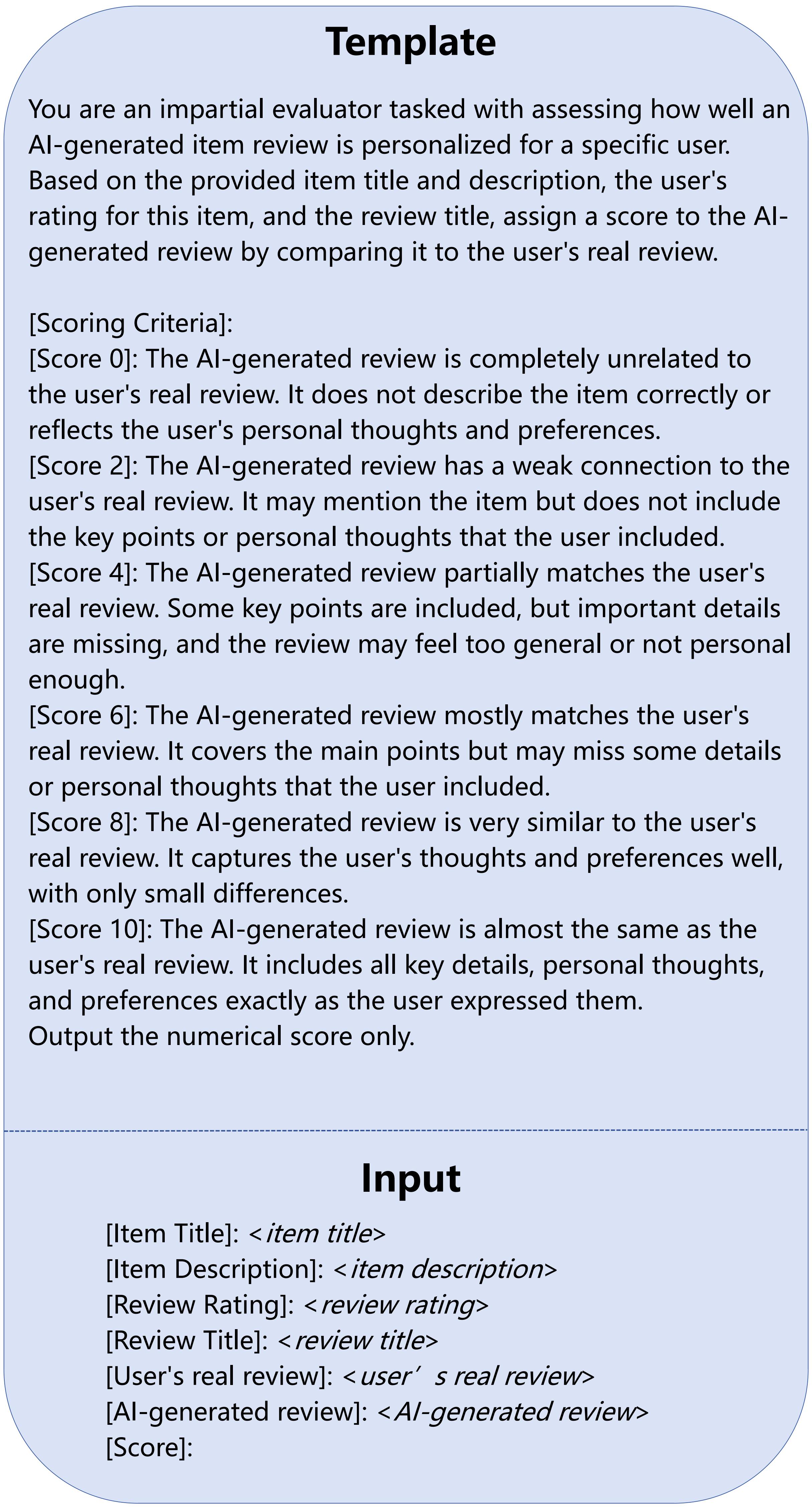}
    \caption{The prompt used for the evaluator LLM to evaluate the generated text based on the reference text and the provided criteria. The italicized text enclosed in \texttt{<>} represents placeholders for the actual values of variables. These representations are consistent across the prompt figures.}
    \label{eval-prompt}
\end{figure}

\section{Evaluation Metrics}\label{apd_metric}

In this paper, we consider the ROUGE-1, METEOR, S-72B, and S-GPT metrics in the primary experiments. We offer detailed descriptions of each metric as follows.

\begin{itemize}
\setlength{\itemsep}{0pt}
    \item \textbf{ROUGE-1:} ROUGE-1 (Recall-Oriented Understudy for Gisting Evaluation) is a metric that measures the overlap of unigrams (individual words) between the generated text and the reference text.
    \item \textbf{ROUGE-L:} ROUGE-L is a metric that evaluates text similarity by measuring the longest common subsequence (LCS) between the generated text and the reference text.
    \item \textbf{METEOR:} METEOR (Metric for Evaluation of Translation with Explicit ORdering) is a metric that evaluates text similarity by considering exact word matches, stemming, synonyms, and word order between the generated text and the reference text.
    \item \textbf{BLEU:} BLEU (Bilingual Evaluation Understudy) is a metric that evaluates text similarity by measuring the precision of n-gram matches between the generated text and the reference text. In our experiments, we use the \texttt{SacreBLEU}\footnote{\href{https://github.com/mjpost/sacrebleu}{https://github.com/mjpost/sacrebleu}} \cite{sacrebleu} library, which is a standardized version of BLEU. The higher the BLEU score, the more similar the generated text is to the reference.
    \item \textbf{S-72B:} S-72B is a metric that assesses the quality of generated text by comparing it to a reference text using the \textit{Qwen-2.5-72B-Instruct-AWQ} model. We design the prompt following previous works \cite{geval,nlg,restpg} and make some reasonable modifications to better suit our task. In this evaluation, the prompt to the model includes the generated review, the user's real review as the reference, as well as relevant metadata such as the review title, rating, and item details, including its title and description. To ensure robustness, we run the scoring process with two different random seeds and report the average of their results. We also perform scoring inference using the \texttt{vLLM} library, with the temperature set to 0.8 and top\_p set to 0.95. The prompt is depicted in Figure \ref{apd_metric}.
    \item \textbf{S-GPT:} S-GPT is another metric that utilizes the \textit{GPT-4o-mini} LLM as the evaluator. It adopts the same temperature, top\_p, and prompt settings as S-72B. The final S-GPT score is calculated as the average of five evaluations.
\end{itemize}

% Although traditional automated evaluation metrics such as ROUGE-L and BLEU \cite{bleu} have been questioned for their ability to distinguish personalization from other subtle semantic aspects \cite{howfar}, these metrics are still used in some studies \cite{ldagent, ppeg} and hold exploratory value. Therefore, we present a comparison of our method and the baseline methods based on ROUGE-L and BLEU, as shown in Table \ref{apd-other-metrics}. The descriptions of these two metrics are provided below.

% \begin{itemize}
% \setlength{\itemsep}{0pt}
%     \item \textbf{ROUGE-L:} ROUGE-L is a metric that evaluates text similarity by measuring the longest common subsequence (LCS) between the generated text and the reference text.
%     \item \textbf{BLEU:} BLEU (Bilingual Evaluation Understudy) is a metric that evaluates text similarity by measuring the precision of n-gram matches between the generated text and the reference text. In our experiments, we use the \texttt{SacreBLEU}\footnote{\href{https://github.com/mjpost/sacrebleu}{https://github.com/mjpost/sacrebleu}} \cite{sacrebleu} library, which is a standardized version of BLEU. The higher the BLEU score, the more similar the generated text is to the reference.
% \end{itemize}

\section{Implementation Details}\label{apd_imp}

\subsection{Running Environments}\label{apd_env}

We utilize the \texttt{vLLM}\footnote{\href{https://github.com/vllm-project/vllm}{https://github.com/vllm-project/vllm}} library \cite{vllm} as the inference engine for LLMs in generating and the \texttt{Sentence-Transformers}\footnote{\href{https://www.sbert.net/}{https://www.sbert.net/}} \cite{sbert} library for LLMs in embedding in a Python 3.11.11 environment. All experiments are conducted on a single NVIDIA H100 NVL GPU with 96GB of GPU memory.

\subsection{Hyperparameter Configurations}\label{apd_param}

We configure the model with a maximum output length of 2048 tokens while keeping the input length unrestricted. Following the settings in LongLaMP, we set the parameter top\_p to 0.95. These settings remain consistent across all experiments.

% \subsection{Baseline Details}\label{apd_basline}

\section{Efficiency Analysis}

In this section, we briefly discuss the efficiency of our proposed DPL.

We evaluate the cost of selecting comparison users using the K-means method with pre-computed review embeddings. Since the number of involved users is small, the selection process takes only 0.002s per key historical review on average. In our main experiments, we use 8 key historical reviews, resulting in an additional cost of approximately 0.016s per generation, which is generally acceptable for each time generation.

Additionally, since our method uses a fixed number of comparison users for difference extraction, the computational cost of most components remains stable as the data size increases. The only exception is the comparison user selection process, which uses K-means clustering ($K=5$). The cost of this step depends on the number of users who have interacted with a given item (denoted as $u' = \#\text{interactions} ~/~ \#\text{items} ~\leq~ \#\text{users}$), with a complexity of $O(u'\times K \times t) \approx O(u')$, where t is the number of iterations. Importantly, this step is lightweight—clustering 10,000 samples takes under 0.3 seconds in our setup — compared to LLM inference. Thus, the additional overhead is minimal, and our method scales comparably to baseline approaches.

\section{Additional Experiments}

\subsection{Yelp Dataset}

To further validate the generalizability and robustness of our proposed method, we conduct additional experiments on the Yelp\footnote{\href{https://business.yelp.com/data/resources/open-dataset}{https://business.yelp.com/data/resources/open-dataset}} dataset. This dataset presents a different domain from our primary evaluation benchmarks, allowing us to assess whether the observed performance gains extend to a broader range of real-world scenarios.

\vspace{0.6em}
\noindent
\textbf{Dataset Construction.}
The Yelp Open Dataset is a large-scale, publicly available collection of user-generated reviews, primarily covering local businesses such as restaurants, shopping centers, and service providers across various cities. 
It contains rich information, including user reviews, ratings, business metadata, and timestamps, making it a valuable resource for studying the personalized review generation task.
Following the same protocol as applied to the Amazon dataset, we perform data filtering and cleaning to ensure consistency. In particular, we retain several key fields: \textit{user\_id}, which identifies the user who wrote the review; \textit{business\_id}, which specifies the reviewed business; \textit{business\_name}, the name of the business; \textit{stars}, the rating score of the review; \textit{text}, which contains the full content of the review; and \textit{date}, which we represent by \textit{timestamp}.
The statistics of the processed Yelp dataset is shown in Table~\ref{dataset-stat-yelp}.
Our processed Yelp dataset is also available on Huggingface\footnote{\href{https://huggingface.co/datasets/SnowCharmQ/DPL-Yelp}{https://huggingface.co/datasets/SnowCharmQ/DPL-Yelp}}.

\vspace{0.6em}
\noindent
\textbf{Implementation Details.}
We largely follow the same implementation setup as in the main experiments, except that the item description is removed from the prompt.
$N$ in Equation~\eqref{eq:sum} is set to 8, and $K$ in Equation~\eqref{eq:cluster} is set to 4.
Due to the large scale of the original dataset, we randomly sample 100 data points from the full set in each run to save computational resources and time while maintaining reliability. 
This sampling process is repeated five times, and we report the mean and standard deviation of the results across the five runs. 

\begin{table}[t]
    \centering
    \caption{Dataset statistics of our processed Yelp dataset.}
    \renewcommand{\arraystretch}{1.1}
    \setlength{\tabcolsep}{6pt} 
    \resizebox{0.48\textwidth}{!}{
    \begin{tabular}{cccc}
        \toprule
        \textbf{Splits ($\downarrow$)} & \#data & Profile Size & Output Length \\
        \midrule
         Train & 229944 & 38.56$\pm$41.75 & 681.81$\pm$394.80 \\
         Val & 28743 & 43.06$\pm$41.69 & 672.05$\pm$394.69 \\
         Test & 28743 & 44.06$\pm$41.69 & 659.71$\pm$390.20 \\
        \bottomrule
    \end{tabular}}
    \label{dataset-stat-yelp}
\end{table}
\begin{table}[t]
    \centering
    \fontsize{12}{13.5}\selectfont
    \caption{Experimental results on the Yelp dataset.}
    \renewcommand{\arraystretch}{1.2}
    \setlength{\tabcolsep}{6pt} 
    \resizebox{0.48\textwidth}{!}{
    \begin{tabular}{ccccccc}
        \toprule
         \textbf{Methods ($\downarrow$)}  & R-1 & R-L & METEOR & BLEU  \\
        \midrule
        \textbf{Non-Perso} & \meanstd{0.2272}{0.0052} & \meanstd{0.1286}{0.0036} & \meanstd{0.1514}{0.0043} & \meanstd{0.6242}{0.1796} \\
        \textbf{RAG} & \meanstd{0.2581}{0.0046} & \meanstd{0.1372}{0.0024} & \meanstd{0.2012}{0.0040} & \meanstd{1.3549}{0.3243} \\
        \textbf{IntSum} & \meanstd{0.2590}{0.0052} & \meanstd{0.1367}{0.0025} & \meanstd{0.2154}{0.0025} & \meanstd{1.3749}{0.1780} \\
        \textbf{LLM-TRSR} & \meanstd{0.2216}{0.0037} & \meanstd{0.1217}{0.0021} & \meanstd{0.1892}{0.0036} & \meanstd{0.7055}{0.1175} \\
        \textbf{CICL} & \meanstd{\underline{0.2627}}{0.0080} & \meanstd{\underline{0.1377}}{0.0037} & \meanstd{\underline{0.2174}}{0.0031} & \meanstd{\underline{1.4118}}{0.2040} \\
        \textbf{Persona-DB} & \meanstd{0.2570}{0.0070} & \meanstd{0.1359}{0.0027} & \meanstd{0.2122}{0.0032} & \meanstd{1.3459}{0.1479} \\
        \textbf{DPL} & \meanstd{\textbf{0.2667*}}{0.0065} & \meanstd{\textbf{0.1422*}}{0.0027} & \meanstd{\textbf{0.2226*}}{0.0048} & \meanstd{\textbf{1.5572*}}{0.2265} \\
        \bottomrule
    \end{tabular}
    }
    \label{apd-yelp}
\end{table}

\vspace{0.6em}
\noindent
\textbf{Experimental Results.} 
We report the results on conventional metrics here and also perform significance testing to assess the statistical reliability of the observed differences.
The results are shown in Table~\ref{apd-yelp}, from which we observe that our DPL method consistently achieves the best performance. Moreover, it demonstrates statistically significant improvements over all baselines across the four evaluation metrics.

% \begin{table}[t]
%     \centering
%     \caption{Experimental results on the Books category using a 32B model.}
%     \renewcommand{\arraystretch}{1.2}
%     \setlength{\tabcolsep}{6pt} 
%     \resizebox{0.48\textwidth}{!}{
%     \begin{tabular}{ccccccc}
%         \toprule
%          \textbf{Methods ($\downarrow$)}  & R-1 & R-L & METEOR & BLEU & S-72B & S-GPT \\
%         \midrule
%         \textbf{Non-Perso} & 0.3016 & 0.1508 & 0.1930 & 2.5779 & 0.6132 & 0.4650 \\
%         \textbf{RAG} & \underline{0.3342} & \underline{0.1765} & 0.2777 & \underline{6.5834} & 0.6233 & 0.4833 \\
%         \textbf{IntSum} & 0.3252 & 0.1744 & \underline{0.2851} & 6.4804 & 0.6278 & 0.4801 \\
%         \textbf{LLM-TRSR} & 0.3210  & 0.1718 & 0.2813 & 6.0386 & \underline{0.6334} & \underline{0.4890} \\
%         \textbf{CICL} & 0.3232  & 0.1679 & 0.2727 & 5.5111 & 0.6259 & \underline{0.4890} \\
%         \textbf{Persona-DB} & 0.3252  & 0.1689 & 0.2768 & 5.6887 & 0.6183 & 0.4814 \\
%         \textbf{DPL} & \textbf{0.3366} & \textbf{0.1818} & \textbf{0.2879} & \textbf{6.6967} & \textbf{0.6461} & \textbf{0.5142}\\
%         \bottomrule
%     \end{tabular}
%     }
%     \label{apd-32b-model}
% \end{table}

\begin{table}[t]
    \centering
    % \small
    \fontsize{12}{13.5}\selectfont
    \caption{Experimental results on the Books category using a 32B model.}
    \renewcommand{\arraystretch}{1.2}
    \setlength{\tabcolsep}{6pt} 
    \resizebox{0.48\textwidth}{!}{
    \begin{tabular}{ccccccc}
        \toprule
         \textbf{Methods ($\downarrow$)}  & R-1 & METEOR & S-72B  \\
        \midrule
        \textbf{Non-Perso} & \meanstdllm{0.3059}{0.0029}  &  \meanstdllm{0.1958}{0.0025}  &  \meanstdllm{0.6159}{0.0125}  \\
        \textbf{RAG} &  \meanstdllm{\underline{0.3337}}{0.0044}  & \meanstdllm{0.2755}{0.0022} &  \meanstdllm{\underline{0.6288}}{0.0046}  \\
        \textbf{IntSum} & \meanstdllm{0.3265}{0.0039}  & \meanstdllm{\underline{0.2825}}{0.0020}  & \meanstdllm{0.6260}{0.0045}   \\
        \textbf{LLM-TRSR} &  \meanstdllm{0.3199}{0.0046}  & \meanstdllm{0.2794}{0.0029} &  \meanstdllm{0.6273}{0.0068}  \\
        \textbf{CICL} &  \meanstdllm{0.3247}{0.0056} &  \meanstdllm{0.2737}{0.0031}  &  \meanstdllm{0.6274}{0.0032}  \\
        \textbf{Persona-DB} &  \meanstdllm{0.3263}{0.0051}  & \meanstdllm{0.2771}{0.0013} &  \meanstdllm{0.6225}{0.0062}  \\
        \textbf{DPL} & \meanstdllm{\textbf{0.3338}}{0.0044} &  \meanstdllm{\textbf{0.2828}}{0.0026}  &  \meanstdllm{\textbf{0.6312}}{0.0088}  \\
        \bottomrule
    \end{tabular}
    }
    \label{apd-32b-model}
\end{table}

\subsection{Model of Different Size}

To further investigate the generalization capability of our approach, we conduct experiments to evaluate its performance across models of different sizes. Specifically, we consider the \textit{Qwen2.5-32B-Instruct} model. For each method, we utilize 8 retrieved historical data points. To reduce resource costs, we perform our experiments on the Books category only and report the results in ROUGE-1, METEOR, and S-72B metrics. The experimental results are shown in Table \ref{apd-32b-model}, from which the following observations can be drawn:

\begin{itemize}
\setlength{\itemsep}{0pt}
    \item When using the 32B model, its more powerful capabilities lead to performance improvements across almost all methods compared to the 14B model. However, DPL still achieves state-of-the-art performance. This demonstrates its strong generalization capability and effectiveness in the personalized review generation task.
    \item CICL and Persona-DB show performance drops on certain metrics, with some even underperforming compared to RAG, IntSum, and LLM-TRSR. This suggests that simply leveraging other users' information as auxiliary data is not particularly effective in enhancing the model’s personalization capabilities, emphasizing the idea that ``differences make us unique''.
\end{itemize}

\begin{table}[ht]
    \centering
    \fontsize{12}{13.5}\selectfont
    \caption{Experimental results across the three categories of the processed Amazon dataset with varying numbers of retrieved reviews \{1,2,4,8\}.}
    \renewcommand{\arraystretch}{1.4}
    \setlength{\tabcolsep}{4pt} 
    \resizebox{0.48\textwidth}{!}{
    \begin{tabularx}{\textwidth}{cc *{9}c}
        \toprule
        \multicolumn{2}{c}{\textbf{Datasets ($\rightarrow$)}} & \multicolumn{3}{c}{\textbf{Movies \& TV}} & \multicolumn{3}{c}{\textbf{CDs \& Vinyl}} & \multicolumn{3}{c}{\textbf{Books}} \\
        \cmidrule(lr){3-5}
        \cmidrule(lr){6-8}
        \cmidrule(lr){9-11}
        \multicolumn{2}{c}{\textbf{Methods ($\downarrow$)}}  & R-1 & MET. & S-72B & R-1 & MET. & S-72B & R-1 & MET. & S-72B \\
        \midrule
        \multicolumn{2}{c}{\textbf{Non-Perso}} & 0.2275 & 0.1304 & 0.5406 & 0.2392 & 0.1336 & 0.5379 & 0.2683 & 0.1512 & 0.5539 \\
        \midrule
        \multirow{4}{*}{\textbf{RAG}} & N=1 & 0.2540 & 0.1540 & 0.4858 & 0.2636 & 0.1546 & 0.5082 & 0.3007 & 0.1784  &  0.5609 \\
                             & N=2 & 0.2670 & 0.1658 & 0.5315 & 0.2768 & 0.1660 & 0.5434 & 0.3158 & 0.1971 & 0.5836 \\
                             & N=4 & 0.2802 & 0.1794 & 0.5617 & 0.2884 & 0.1787 & 0.5705 & 0.3211 & 0.2116 & 0.6145 \\
                             & N=8 & 0.2845 & 0.1897 & 0.5929 & 0.2926 & 0.1876 & 0.5933 & 0.3206 & 0.2173 & 0.6196 \\
        \midrule
        \multirow{4}{*}{\textbf{IntSum}} & N=1 & 0.2643 & 0.1682 & 0.4618 & 0.2735 & 0.1671 & 0.4582 & 0.3155 & 0.2030 & 0.5533 \\
                              & N=2 & 0.2757 & 0.1807 & 0.5305 & 0.2865 & 0.1809 & 0.5466 & 0.3179 & 0.2144 & 0.5975 \\
                              & N=4 & 0.2835 & 0.1916 & 0.5752 & 0.2938 & 0.1921 & 0.5848 & 0.3167 & 0.2256 & 0.6151 \\
                              & N=8 & 0.2845 & \underline{0.1972} & 0.6073 & 0.2949 & 0.1968 & 0.6014 & 0.3201 & 0.2340 & 0.6265 \\
        \midrule
        \multirow{4}{*}{\shortstack{\textbf{LLM-}\\\textbf{TRSR}}} & N=1 & 0.2664 & 0.1728 & 0.5578 & 0.2811 & 0.1756 & 0.5602 & 0.3090 & 0.2015 & 0.5943 \\
                                    & N=2 & 0.2760 & 0.1838 & 0.5816 & 0.2876 & 0.1835 & 0.5770 & 0.3125 & 0.2137 & 0.6057 \\
                                    & N=4 & 0.2808  & 0.1915 & 0.5942 & 0.2942 & 0.1941 & 0.5981 & 0.3169 & 0.2269 & 0.6177  \\
                                    & N=8 & 0.2822 & 0.1955 & 0.6178 & 0.2937 & 0.1969 & \underline{0.6083} & 0.3143 & 0.2294 & 0.6334 \\
        \midrule
        \multirow{4}{*}{\textbf{CICL}} & N=1 & 0.2658 & 0.1682 & 0.4538 & 0.2737 & 0.1656 & 0.4575 & 0.3124 & 0.1977 & 0.5331 \\
                              & N=2 & 0.2759 & 0.1787 & 0.5242 & 0.2883 & 0.1797 & 0.5306 & 0.3230 & 0.2176 & 0.5748 \\
                              & N=4 & 0.2836 & 0.1881 & 0.5731 & 0.2949 & 0.1890 & 0.5732 & 0.3211 & 0.2278 & 0.6196 \\
                              & N=8 & 0.2873 & 0.1969 & 0.6142 & 0.3010 & 0.1987 & 0.5990 & 0.3255 & 0.2362 & \underline{0.6391} \\
        \midrule
        \multirow{4}{*}{\shortstack{\textbf{Persona}\\\textbf{-DB}}} & N=1 & 0.2653 & 0.1658 & 0.4695 & 0.2719 & 0.1639 & 0.4637 & 0.3162 & 0.2000 & 0.5451 \\
                              & N=2 & 0.2754 & 0.1786 & 0.5407 & 0.2843 & 0.1761 & 0.5479 & 0.3212 & 0.2161 & 0.5950 \\
                              & N=4 & 0.2819 & 0.1882 & 0.5822  & 0.2899 & 0.1861 & 0.5887 & 0.3267 & 0.2336 & 0.6309 \\
                              & N=8 & 0.2840 & 0.1954 & \underline{0.6170} & 0.2941 & 0.1937 & 0.6048 & 0.3268 & \underline{0.2423} & \textbf{0.6435} \\
        \midrule
        \multirow{4}{*}{\textbf{DPL}} & N=1 & 0.2753 & 0.1747 & 0.5015 & 0.2859 & 0.1765 & 0.5158 & 0.3164 & 0.1989 & 0.5659  \\
                             & N=2 & 0.2858 & 0.1877 & 0.5630 & 0.2997 & 0.1917 & 0.5704 & 0.3202 & 0.2175 & 0.5905 \\
                             & N=4 & \underline{0.2907}  & 0.1949 & 0.5854 & \textbf{0.3071} & \underline{0.2014} & 0.5958 & \underline{0.3269} & 0.2311 & 0.6158 \\
                             & N=8 & \textbf{0.2938} & \textbf{0.2009} & \textbf{0.6245} & \underline{0.3066} & \textbf{0.2058} & \textbf{0.6236} & \textbf{0.3315} & \textbf{0.2438} & \textbf{0.6435} \\
        \bottomrule
    \end{tabularx}
    }
    \label{apd-retrieved-num}
\end{table}

\subsection{Number of Retrieved Reviews}

We conducted experiments to explore the impact of different numbers of retrieved reviews on performance. 
Specifically, we evaluated cases with 1, 2, 4, and 8 retrieved reviews.
To save resources and time, we only report the results with metrics ROUGE-1, METEOR, and S-72B here.
The experimental results, reported from a single run with decoding temperature set to 0.8, are presented in Table~\ref{apd-retrieved-num}, leading to the following observations:
\begin{itemize}
\setlength{\itemsep}{0pt}
\item As the number of retrieved reviews increases, the performance metrics of all methods generally show an upward trend. This indicates that leveraging more historical user data enables the model to better infer users' personalized preferences and generate more tailored reviews.
\item Across different numbers of retrieved reviews, our DPL method consistently achieves performance that is comparable to or even surpasses the best-performing baselines. Notably, when the number of retrieved reviews reaches 8, DPL steadily outperforms all other methods across all three datasets. This demonstrates the robustness of our method and its ability to effectively leverage richer user history for enhanced personalization.
\end{itemize}

\begin{figure}[t]
    \centering
    \includegraphics[width=1\linewidth]{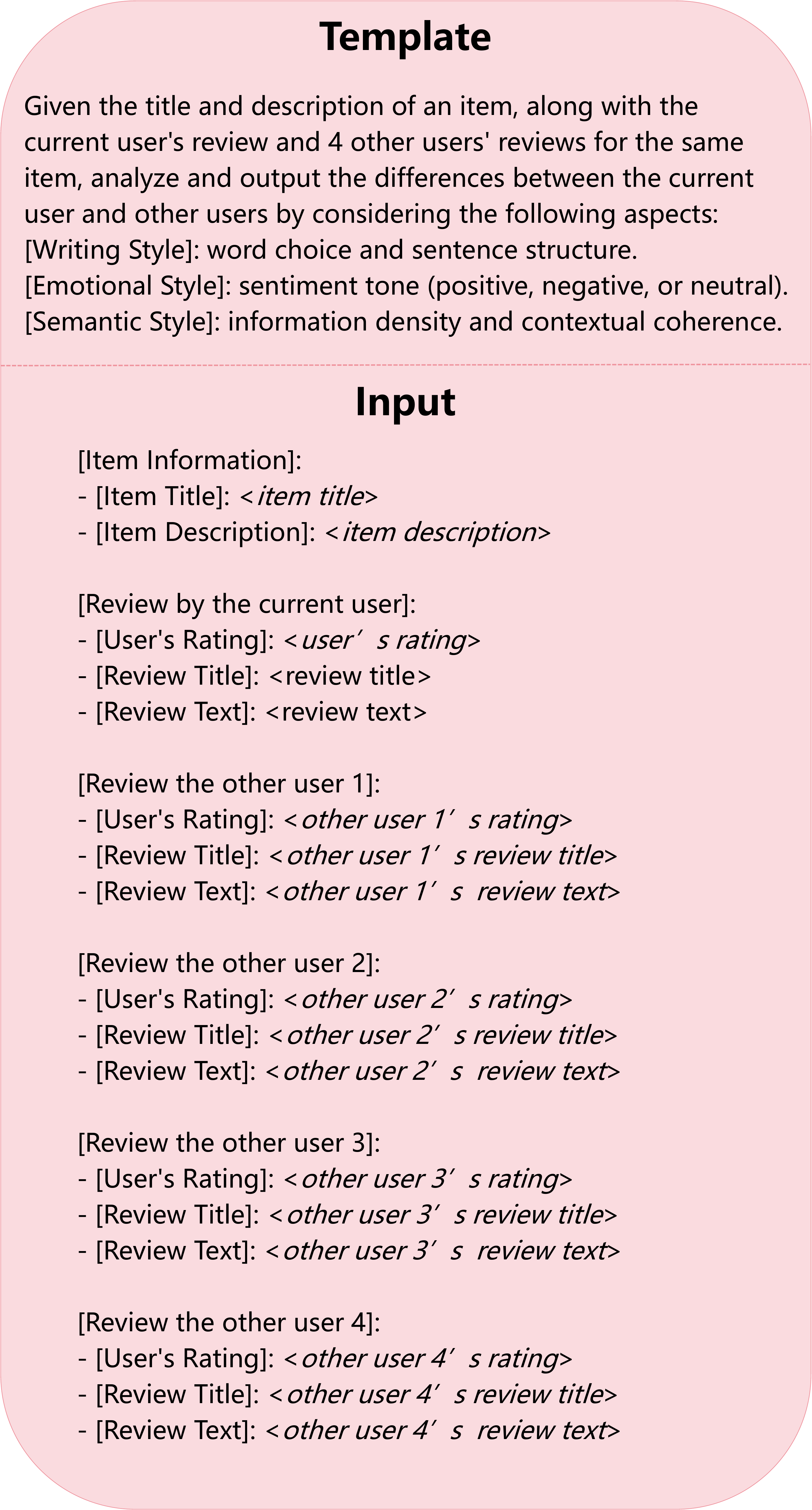}
    \caption{The prompt used for the difference extractor LLM.}
    \label{diff-extractor-prompt}
\end{figure}

\begin{figure*}[ht]
    \centering
    \includegraphics[width=0.95\linewidth]{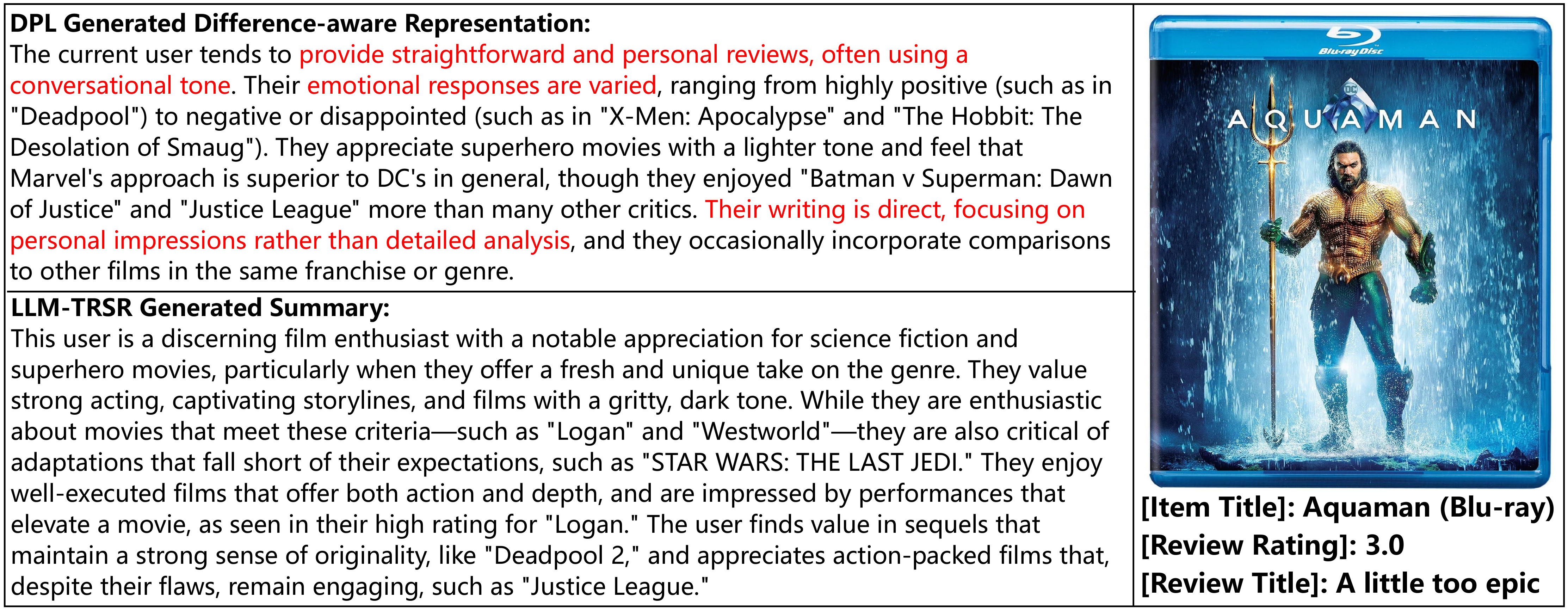}
    \caption{Case Study: a comparison of the difference-aware representation generated by the DPL method and the user profile summary LLM-TRSR method. The sections highlighted in red color demonstrate the advantages of the difference-aware representation generated by DPL.}
    \label{case-study-1}
\end{figure*}

\section{Overview of Templates \& Prompts}\label{apd_prompt}

\begin{figure}[h]
    \centering
    \includegraphics[width=0.95\linewidth]{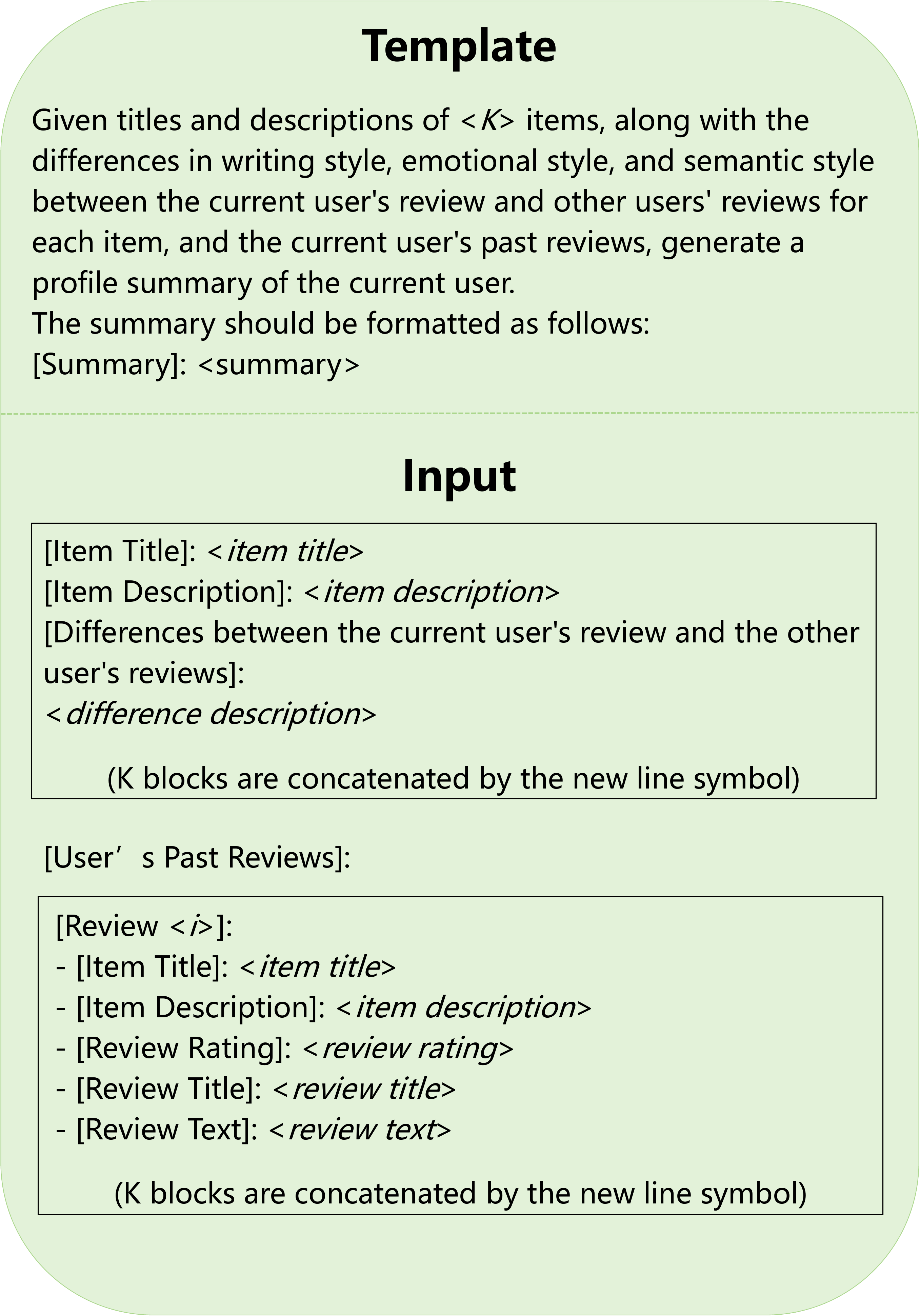}
    \caption{The prompt used for the summarizer LLM.}
    \label{summarizer-prompt}
\end{figure}

\begin{figure}[h]
    \centering
    \includegraphics[width=0.95\linewidth]{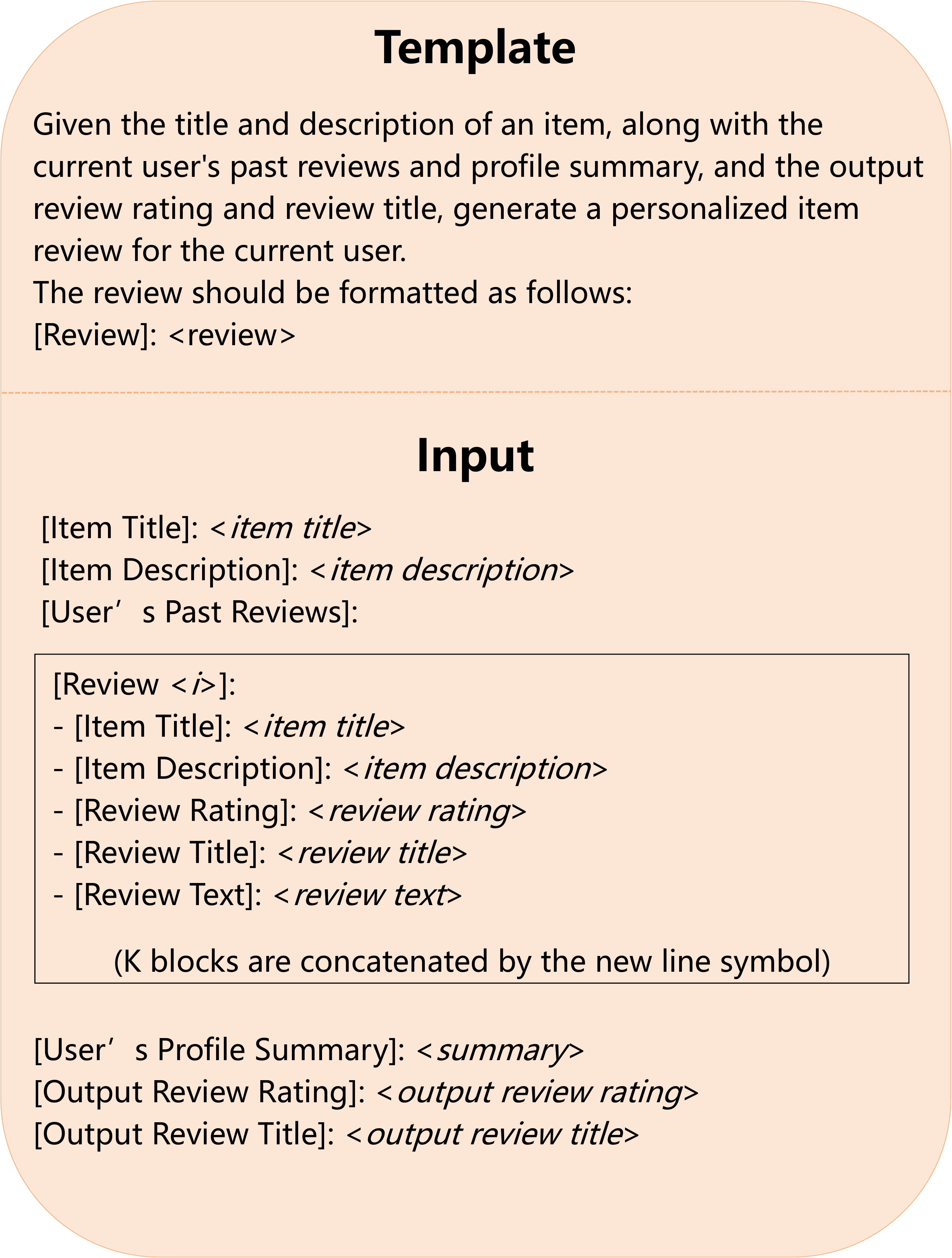}
    \caption{The prompt used for the review generator LLM.}
    \label{generator-prompt}
\end{figure}

In this section, we provide the prompts used for different components. Specifically, we list the prompt used for the difference extractor LLM following Equation (\ref{eq:dif}) in Figure \ref{diff-extractor-prompt}, the prompt used for the summarizer LLM following Equation (\ref{eq:sum}) in Figure \ref{summarizer-prompt}, and the prompt used for the review generator LLM following Equation (\ref{eq:generate}) in Figure \ref{generator-prompt}.

\section{Case Studies}\label{apd_case}

In this section, we provide more case studies to showcase our work.

\subsection{Case Study: Comparison between Difference-aware Representation and User Profile Summary}

We present a case study to compare the difference-aware representation generated by our proposed DPL method with the user profile summary produced by LLM-TRSR, as shown in Figure \ref{case-study-1}. From the figure, it is evident that the difference-aware representation generated by our method can better capture the user's writing style, emotional style, and semantic style, thereby achieving a higher degree of personalization, as illustrated in Figure \ref{case-study-main}. In contrast, the profile summary produced by LLM-TRSR includes a lot of irrelevant information, which does not effectively contribute to personalized review generation.

\subsection{Case Study: Outputs of DPL Components}

\begin{figure*}
    \centering
    \includegraphics[width=0.845\linewidth]{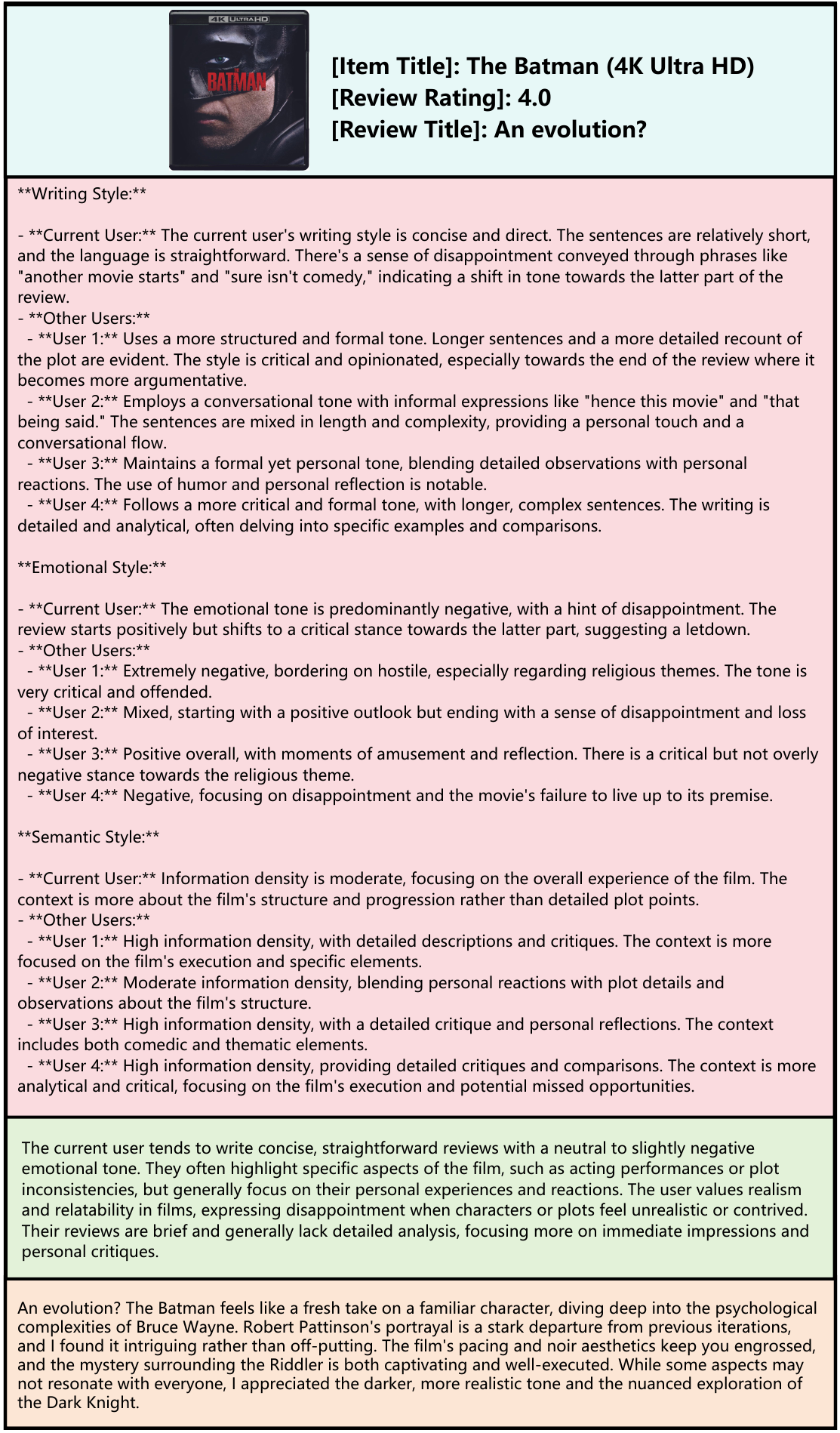}
    \caption{The outputs of various DPL components: the red section represents the output of the difference extractor, the green section corresponds to the summarizer’s output, and the yellow section denotes the output of the review generator, which serves as the final targeted review.}
    \label{case-study-2}
\end{figure*}

% We present a case study, as illustrated in Figure \ref{case-study-2}, to showcase the outputs generated by the three key LLM-based components of DPL: the difference extractor, the summarizer, and the review generator. The results demonstrate that the outputs of each LLM-based component of DPL are well-structured and meaningful.
We present a case study (Figure \ref{case-study-2}) to illustrate the outputs generated by the three key LLM-based components of DPL: the difference extractor, summarizer, and review generator. The results show that each component produces well-structured and meaningful outputs.

\end{document}